\documentclass{article} 
\usepackage{iclr2025_conference,times}

\usepackage{amsmath,amsfonts,bm}

\def\eqref#1{equation~\ref{#1}}

\def\1{\bm{1}}

\DeclareMathAlphabet{\mathsfit}{\encodingdefault}{\sfdefault}{m}{sl}
\SetMathAlphabet{\mathsfit}{bold}{\encodingdefault}{\sfdefault}{bx}{n}

\DeclareMathOperator*{\argmin}{arg\,min}

\usepackage{hyperref}
\usepackage{url}

\usepackage{pgfplots}
\usepackage{ dsfont }
\usepackage{placeins} 
\usepgfplotslibrary{groupplots}
\usepackage[normalem]{ulem}
\usepackage{adjustbox}
\usepgfplotslibrary{fillbetween}
\usepackage[ruled,vlined]{algorithm2e}
\usepackage{pifont}
\usetikzlibrary{backgrounds}
\usepackage{wrapfig}
\usepackage{booktabs}
\usepackage{tikz}

\usepackage{caption}
\usepackage{subcaption}
\usepackage{graphicx}
\usepackage{contour}

\newcommand{\cmark}{\ding{51}}
\newcommand{\xmark}{\ding{55}}

\definecolor{baselinecolor}{HTML}{00008b}
\definecolor{freshcolor}{HTML}{008000}
\definecolor{relucolor}{HTML}{1e90ff}
\definecolor{sirencolor}{HTML}{ff8c00}
\definecolor{color1}{HTML}{2f4f4f} 
\definecolor{color2}{HTML}{00ff00}
\definecolor{color3}{HTML}{ff4444}  
\definecolor{color4}{HTML}{ff69b4}
\definecolor{color5}{HTML}{00ffff}
\definecolor{colorfig2line}{HTML}{BD0000}

\def\x{\mathbf{x}}
\def\w{\mathbf{W}}
\def\wrow{\mathbf{w}_i}

\def\our{FreSh}
\def\siren{{\em Siren}}
\def\ffeatures{{\em Fourier}}
\def\finer{{\em Finer}}
\def\wire{{\em Wire}}

\def\chest{Chest X-Ray}
\def\ffhqcropped{FFHQ-1024}
\def\ffhq{FFHQ-wild}
\def\kodak{Kodak}
\def\art{Wiki Art}

\def\spectrum{\mathcal{S}}

\usepackage{amsmath}
\usepackage{cleveref}

\newcommand{\real}[0]{\mathbb{R}}

\title{FreSh: Frequency Shifting for Accelerated Neural Representation Learning}

\author{Adam Kania$^1$, Marko Mihajlovic$^2$, Sergey Prokudin$^{2,3}$, Jacek Tabor$^1$, Przemysław Spurek$^{1,4}$ \\
Jagiellonian University$^1$; ETH Zurich$^2$; Balgrist University Hospital$^3$; IDEAS NCBR$^4$ \\
}

\iclrfinalcopy 
\begin{document}

\maketitle

\begin{abstract}
Implicit Neural Representations (INRs) have recently gained attention as a powerful approach for continuously representing signals such as images, videos, and 3D shapes using multilayer perceptrons (MLPs). However, MLPs are known to exhibit a low-frequency bias, limiting their ability to capture high-frequency details accurately. This limitation is typically addressed by incorporating high-frequency input embeddings or specialized activation layers. In this work, we demonstrate that these embeddings and activations are often configured with hyperparameters that perform well on average but are suboptimal for specific input signals under consideration, necessitating a costly grid search to identify optimal settings. Our key observation is that the initial frequency spectrum of an \textit{untrained} model's output correlates strongly with the model's eventual performance on a given target signal. Leveraging this insight, we propose \textit{frequency shifting} (or \our{}), a method that selects embedding hyperparameters to align the frequency spectrum of the model’s initial output with that of the target signal. We show that this simple initialization technique improves performance across various neural representation methods and tasks, achieving results comparable to extensive hyperparameter sweeps but with only marginal computational overhead compared to training a single model with default hyperparameters.
\end{abstract}

\section{Introduction}

Implicit Neural Representations (INRs) are advancing computer graphics research by integrating classical algorithms with continuous signal representations. 
They have been successfully applied in signal representation and inverse problems, with notable applications in neural rendering, compression, and 2D and 3D signal reconstruction \citep{xie2022neural}. 

INRs primarily rely on multilayer perceptrons (MLPs), making them susceptible to {\it spectral bias}, which refers to the slower convergence of MLPs when approximating high-frequency components of the target signal \citep{SpectralBiasOfNN}. Although spectral bias can benefit generalization \citep{ronen2019convergence}, it can also hinder performance \citep{WalshHadamardRegularizer, SpectralBiasOfNN}, especially in scenarios that require high precision of signal reconstruction, such as the training of INRs. 
This led to the development of numerous architectures aimed at overcoming {\it spectral bias} and its resulting capacity constraints by increasing the frequencies present in the input signal at the first layer of the model (embedding layer), e.g., through frequency-changing activation functions \citep{SIREN,liu2024finer, FourierFeatures} or auxiliary data structures \citep{chan2022triplane,muller2022instant}.

\begin{figure}[htbp]
    \vspace{-2mm}
    
    \includegraphics[width=\linewidth]{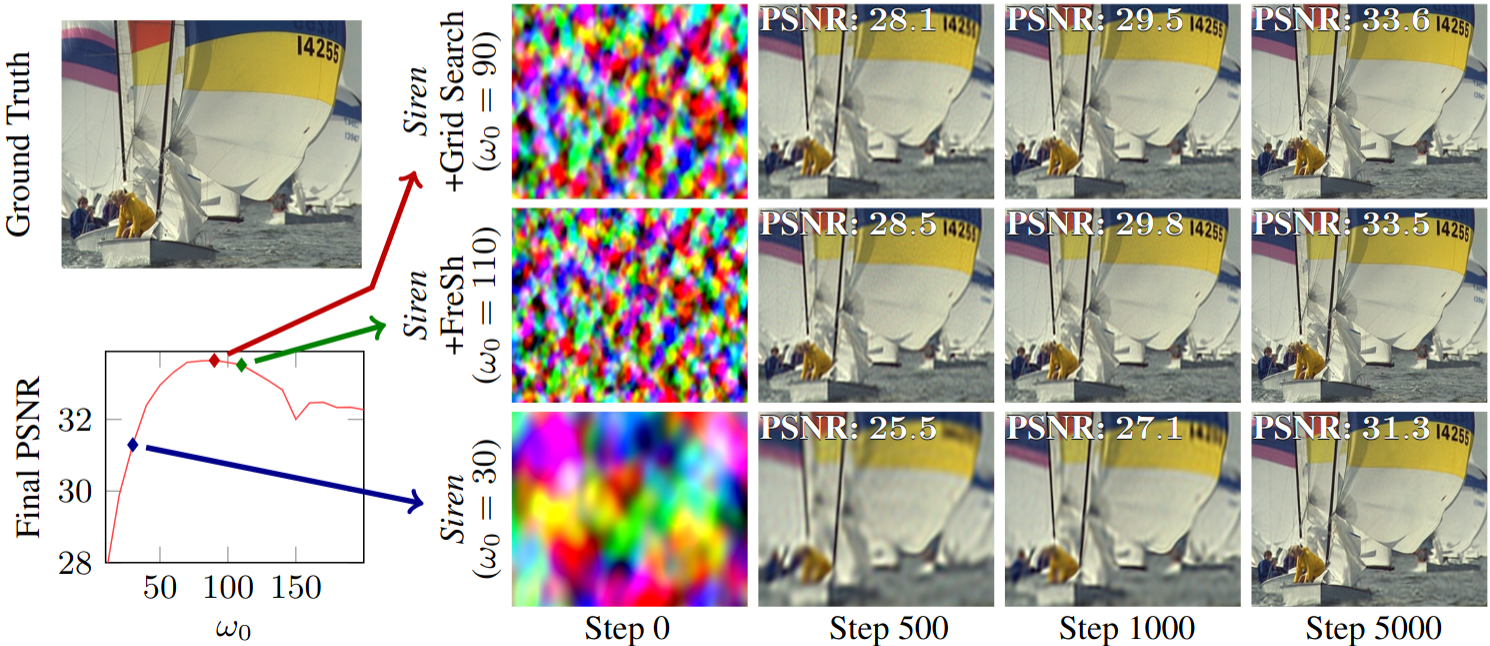} 
    \centering
    
    \caption{
    The configuration of embeddings is crucial for the convergence speed. We train \siren{} with various embedding configurations ($\omega_0 \in [10, 200]$) for 5k steps on a \kodak{} image (top-left). The best grid-search found model ($\omega_0=90$), the \our{} configuration ($\omega_0=110$) and the baseline ($\omega_0=30$) are marked with diamonds (bottom-left). The optimal and \our{} configurations (top and middle rows) lead to sharper details, such as the number on the sail, compared with the baseline (bottom row). Even though \siren{} uses a frequency embedding, the baseline is blurry due to low frequency bias. Note how the sizes of uniformly colored areas in the output at step 0 indicate the size of image features the network can easily learn - this observation is pivotal for \our{}.}
    \label{fig:color_steps}
    \vspace{-4mm}
\end{figure}

The frequency of such embedding layers is typically controlled by hyperparameters whose configuration can significantly affect performance (see \Cref{fig:color_steps}). In \cref{sec:motivation} we show that default hyperparameter values can hinder performance and lead to blurry reconstructions. Improving performance is possible by using parameters sweeps,

however, optimizing embedding parameters by training multiple models introduces significant overhead and is not feasible in practice. 

The high computational cost of performing a parameter sweep comes from training each model. Instead of training, we approximate model performance using the Fourier transform and the Wasserstein distance \citep{wasserstein1969markov}, significantly reducing the computational costs (see \Cref{tab:image_fitting_times}). Our method, dubbed \our{} (\textbf{Fre}quency \textbf{Sh}ifting for Accelerated Neural Representation Learning), selects the embedding configuration where the frequency distribution of the model's output is close to that of the target signal. This shift in the model's frequency distribution results in better signal modeling (see \Cref{fig:color_steps}). We validate our approach experimentally, demonstrating improved quality in representation tasks such as image and video overfitting and in an inverse problem, specifically 3D shape modeling with NeRF. We show that these improvements result from a more accurate approximation of all frequencies (see \Cref{fig:img_residuals}).

In existing INR studies \citep{SIREN,FourierFeatures, WIRE,liu2024finer,muller2022instant}, new architectures are often introduced with minimal attention to simplifying the costly process of hyperparameter selection, even though it is crucial for achieving the best performance. Our framework addresses this gap by leveraging frequency information to guide the selection process for embeddings of various INR models. Our key contributions are:

\begin{itemize}
    
    \item {We develop a technique for comparing frequency contents of images based on the Discrete Fourier Transform and the Wasserstein distance.}
    \item We introduce \our{} (\textbf{Fre}quency \textbf{Sh}ifting for Accelerated Neural Representation Learning) - a model agnostic method for configuring coordinate embeddings 
    for better performance, that can be easily applied using a provided script \footnote{We release code for \our{} at \url{https://github.com/gmum/FreSh/}
    }. \our{} works by adjusting the capacity of an INR to model all target signal's frequencies.

    \item {We achieve state-of-the-art results in image and video approximation, as well as 3D shape reconstruction (NeRF), using a fraction of the compute of a conventional grid search}.
\end{itemize}

\section{Related Work}
\label{sec:related}

\textbf{INRs} are neural models used for signal representation that received considerable attention in research \citep{tewari2022advances} and have been applied in various domains, including representation of images \citep{klocek2019hypernetwork}, videos \citep{chen2022videoinr}, 
and 3D shapes \citep{park2019deepsdf}. Notable applications include 3D shape reconstruction \citep{Nerf}, robotics \citep{wang2021nerf, lin2021barf}, and compression \citep{lu2021compressive, takikawa2021neural}. INR architectures are often simple, consisting of a single MLP, with improvements focusing on embedding layers \mbox{\citep{chen2022tensorf, muller2022instant}}, activations \citep{SIREN}, rendering techniques \citep{barron2021mip, barron2023zip}, and regularization methods \citep{yang2023freenerf}.

\textbf{Spectral bias} is a phenomenon observed in MLPs describing their preference for learning low-frequency functions and ignoring high-frequency noise \citep{SpectralBiasOfNN, ronen2019convergence, DeepLearningFourierAnalysis}, which helps explain the remarkable generalization properties of deep models. Nevertheless, this low-frequency bias hurts model performance when high-frequency components of the signal are informative \citep{WalshHadamardRegularizer}. A common way of overcoming this bias in INRs is by introducing high-frequency embeddings \citep{SIREN,Nerf,muller2022instant,liu2024finer} that change the space over which the MLP operates, thus modifying the frequencies of the target function. However, the effectiveness of such approaches is limited as optimal hyperparameter configurations have to be found by trial and error for each target signal.

\textbf{Positional encodings} are a broad class of functions that map coordinates into a high dimensional space through a function of adjustable frequency. Their particularly prominent use is in Transformers \citep{vaswani2017attention}, but they are also crucial for INRs. One of their first applications was NeRF \citep{Nerf}, a model for 3D scene reconstruction from a set of posed 2D images. 
NeRF embeddings are not stable in sparse settings due to its usage of very high frequencies \citep{yang2023freenerf}, and its axis-alignment makes reconstruction quality rotation-dependent \citep{FourierFeatures}. Despite its drawbacks, it is frequently used \citep{barron2021mip, pumarola2021d, barron2023zip}, which could be due to its low sensitivity to hyperparameters compared to alternatives. These alternatives include various activation functions \citep{SIREN, WIRE} and a direction-invariant version of NeRF using Fourier features \citep{FourierFeatures}.

\textbf{Activation functions} are a popular method of improving the capacity of INRs, where the first layer acts as a positional embedding. A well-known example is \siren{} \citep{SIREN}, which uses a sine activation to increase signal frequencies in the model's first layer. 
Architectures that generalize this approach
utilize the Gabor wavelet \citep{WIRE}, non-periodic functions \citep{ramasinghe2022beyond}, and variable-periodic functions \citep{liu2024finer}. In this work, we particularly focus on \siren{} due to its popularity, but we also address other activations.

\textbf{Auxiliary data structures} are used in neural scene representation to associate fragments of the scene with a feature vector of trainable parameters, trading a larger memory footprint for smaller computational costs. However, due to the extremely small MLPs used, these approaches can lose some of global reasoning and implicit regularization \citep{neyshabur2014search, goodfellow2016deep} capabilities of neural models. As directly storing a fine grid of features would be prohibitively expensive, practical approaches use low-rank approximations \citep{chen2022tensorf}, 2D feature maps \citep{chan2022triplane} and hash tables \citep{muller2022instant} to reduce the memory cost. In such settings spectral bias is not avoided, and the resolution of the voxel grid needs to be tuned for each scene. 
 Although our main goal is to improve pure neural network-based solutions, we also verify on the recent architecture from \citet{muller2022instant} that \our{} is applicable to grid-based approaches. 

\textbf{ResFields} is a novel framework for INRs that improves their capacity for representing complex signals by incorporating temporal residual layers into MLPs \citep{mihajlovic2023resfields}. By modifying network weights with a time-dependent component represented as a factorized matrix, it increases the performance with only a small impact on parameter count and inference speed. We use ResFields to improve current state-of-the-art results on video representation.

\textbf{Initialization schemes for INRs} have been extensively studied to accelerate training convergence. IGR~\citep{IGR} introduced an implicit geometric initialization to speed up learning 3D shapes, while others~\citep{rajeswaran2019meta,sitzmann2020metasdf,wang2021metaavatar,tancik2021learned} leveraged data-driven meta-learning approaches~\citep{finn2017model,nichol2018first} for learning implicit fields. In contrast to these methods, which rely on computationally expensive pre-training or hand-crafted priors, our approach avoids such requirements, offering a more efficient alternative.

\textbf{Regularization strategies} can be applied to stabilize the training of NeRF-based models, especially in sparse settings \citep{yang2023freenerf}. As our interest is in hyperparameter selection, we do not test the effects of regularization on final results. Moreover, some regularization methods \citep{yang2023freenerf} were developed only for NeRF-like embeddings and would have to be generalized for a fair comparison. It is also worth noting that the Wasserstein distance has been already applied to improve INRs \citep{ramasinghe2024improving} through regularization, which is different from our application.

\section{Motivation}
\label{sec:motivation}

In this section, we show that the training of an INR highly depends on configuring the embedding in a way that aligns its frequencies with the target signal. This leaves practitioners with two options, either using a suboptimal, default embedding configuration or finding a well performing configuration through a costly parameter sweep.

\begin{table}[t!]
    \centering
    \caption{Comparison between the default, optimal, and \our{} configurations of \siren{} during 15,000-step training on various datasets, using one image per dataset. We provide the PSNR scores and the total training time, which includes $\omega_0$ selection. The default value of $\omega_0$ leads to suboptimal results, but improvements can be achieved by using an optimal configuration found through a hyperparameter sweep ($\omega_0 \in \{30, 40, ..., 140\}$). This process involves training multiple models, making fixing $\omega_0$ a common strategy. \our{} achieves similar improvements to grid search while being an order of magnitude faster and comparable in time with a plain \siren{}. Results are averaged over three seeds. The grid search was performed once, and the best configuration was re-evaluated using different seeds. }
    \label{tab:sweep_vs_default_and_time}
     \begin{tabular}{c|cccc|c}
            PSNR $\uparrow$  &   \multicolumn{4}{c|}{ Step} & Training \\
               & 500 & 1000 & 5000 & 15000 &  time (h) $\downarrow$\\
            \midrule
            { \siren{} $(\omega_0 = 30)$} & $25.26$ $\mathsmaller{\pm  0.03}$ & $26.10$ $\mathsmaller{\pm  0.04}$ & $29.13$ $\mathsmaller{\pm  0.06}$ & $31.18$ $\mathsmaller{\pm  0.02}$  &   $2.66 \mathsmaller {\pm >0.01}$ \\
            { \siren{} + Grid Search} & $27.36$ $\mathsmaller{\pm  0.06}$ & $28.39$ $\mathsmaller{\pm  0.03}$ & $30.51$ $\mathsmaller{\pm  0.03}$ & $32.02$ $\mathsmaller{\pm  0.02}$  & $32.69 \mathsmaller {\pm 0.16}$ \\
             { \siren{} + \our{}} & $27.11$ $\mathsmaller{\pm  0.03}$ & $28.16$ $\mathsmaller{\pm  0.02}$ & $30.32$ $\mathsmaller{\pm  0.03}$ & $31.81$ $\mathsmaller{\pm  0.01}$ & $2.82 \mathsmaller{\pm 0.01}$ \\
        \bottomrule    
\end{tabular}
\end{table}

We illustrate the impact of proper hyperparameter selection in an image representation task in \Cref{fig:color_steps} using the \siren{} model, which uses an input-scaling parameter $\omega_0$ to increase embedding frequencies. We compare a default, unaligned with the target signal configuration of \siren{} ($\omega_0=30$) to an aligned configuration ($\omega_0=90$) selected through a parameter sweep over $\omega_0 \in \{30,\dots,140\}$. The aligned configuration speeds up training by employing frequencies three times greater than the baseline, achieving sharp details while the baseline is blurry. Note how the sizes of uniformly colored areas in the output at step 0 indicate the size of features the network can easily learn - this observation is pivotal for \our{}. We perform a similar investigation on 5 images, each from a different dataset (see \cref{sec:experiments}), reporting the results in \Cref{tab:sweep_vs_default_and_time}. We find that optimizing $\omega_0$ always improves the baseline results, with specific best values of $\omega_0$ depending on the target signal. The failure of SGD to optimize the embedding layer (see \Cref{app:sgd}) necessitates hyperparameter sweeps, as a one-size-fits-all solution will inevitably be suboptimal for some target signals. Our goal is to perform this search and enhance INR performance while avoiding the high computational cost of repeated model re-training.

\section{\our{}}
\label{sec:method_desc}

In this section, we discuss how to compare the frequency contents of images and introduce \our{}, a computationally efficient method for initializing frequency embeddings that biases the model towards the frequencies present in the target signal.

\subsection{Preliminaries}
\label{sec:preliminaries}

This section discusses important theoretical concepts relevant to our study. INR architectures are discussed in the Appendix, with the exception of the \siren{} model \citep{SIREN}, which we use as a high-level example to illustrate how similar approaches work. We provide a list of embedding hyperparameters from each model used in our study in \Cref{tab:model_parameters}.

\textbf{\siren{}} \citep{SIREN} addresses spectral bias by mapping its inputs, $\mathbf{x} \in \real^d$, through a high-frequency embedding, given as:
\begin{equation}\label{eq:siren_embedding}
    \gamma_S(\x) = \sin(\omega_0W\x + \mathbf{b}),
\end{equation}
where $W\sim \mathcal{U}[-\frac 1 d, \frac 1 d]$ are the weights of the layer and $\mathbf{b}$ is bias. The scaling parameter $\omega_0$ controls the frequency magnitudes of this embedding and the authors \citep{SIREN} recommend setting $\omega_0=30$, but adjustments are needed to reach optimal performance \citep{WIRE}. \siren{} was specifically designed so that the first layer is the sole contributor to the frequency increase inside the model, which results in the frequency distribution (spectrum) of this model being very concise.

\begin{figure}[tbp]
    \centering
    \centering
    \input{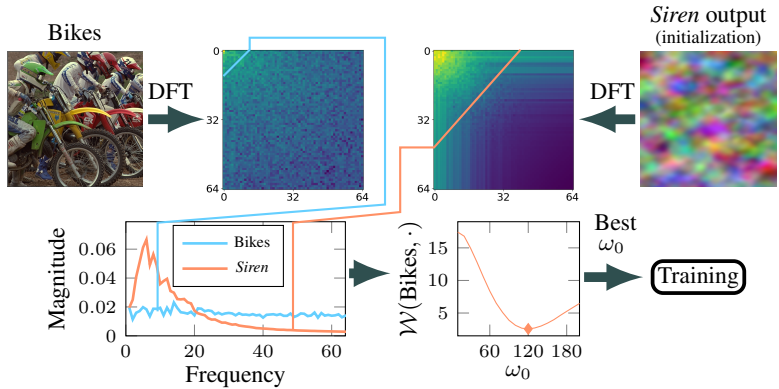}
    \caption{Example workflow of \our{} when applied to \siren{} and a high-frequency \kodak{} image. First, the image and outputs from various model configurations undergo a Discrete Fourier Transform (DFT). The Fourier coefficients of the same degree are then summed to produce the image spectrum (bottom-left). The model spectra are compared with the dataset spectrum using the Wasserstein distance $\mathcal{W}$ (bottom-middle), with only the configuration at the global minimum, highlighted by a diamond, used for training. Note that the Wasserstein distance follows a smooth trend with a distinct global minimum, indicating stable and predictable behavior.
    }
    \label{fig:fresh_workflow_and_wasserstein}
    \vspace{-5mm}
\end{figure}

\begin{wraptable}{r}{0.43\linewidth}
  \vspace{-4.6mm}
  \caption{Embedding hyperparameters of architectures used in our study. In all models increasing the hyperparameter value results in higher embedding frequencies.}
  \label{tab:model_parameters}
  \centering
        \begin{tabular}{@{\hspace{1mm}}l@{\hspace{2mm}}l@{\hspace{2mm}}l@{\hspace{1mm}}}
        \toprule
        ~     & \small Frequency & ~ \\ 
        \small Model & \small Parameter & \small Description \\ 
        \midrule
        \small \ffeatures{} & $\sigma$  & \small Weight variance \\
        \small \siren{} & $\omega_0$  & \small Scaling parameter \\
        \small \finer{} & $\omega$, $k$  & \small Scaling parameters \\
        Hashgrid & \small $\text{N}_\text{max}$ & \small  Grid resolution \\
        \bottomrule
    \end{tabular}
    
\end{wraptable}
\textbf{Other models} adopt a similar approach to \siren{} and control embedding frequencies through a hyperparameter, with higher values corresponding to higher frequencies. Notable models that employ this strategy include NeRF \citep{Nerf}, \ffeatures{} \citep{FourierFeatures}, \finer{} \citep{liu2024finer}, and Hashgrid \citep{muller2022instant}. While most models use embedding frequencies that are similar to what is present in the dataset and not excessively high, NeRF employs frequencies that often surpass those found in the dataset, which makes it incompatible with our comparison-based method. 
Similarly, \wire{} \citep{WIRE} uses very high frequencies and increases frequencies at each hidden layer, not just at the embedding layer. This makes it incompatible with \our{}. As such, we provide results for NeRF and \wire{} as a reference, but do not optimize them with \our{}. The approach of \finer{} differs from other models by employing two hyperparameters, though they largely serve the same purpose, raising questions about the necessity of both. We explore two scenarios for \finer{}: in each scenario, we optimize only one of the parameters.

\textbf{Discrete Fourier Transform} (DFT) of an image $A \in \real^{C \times N \times N}$ describes its frequencies based on both their magnitude and direction. For example, $\sin(x_0)$ and $\sin (x_1)$ represent the same frequency magnitude but different directions. We denote the DFT of the $c$-th channel of $A$ as $\mathcal{F}(A_c) \in \real^{N \times N}$, where the element $\mathcal{F}_{j,k}(A_c)$ is defined as
\begin{equation}
    \mathcal{F}_{j,k}(A_c) = \sum_{m=0}^{N-1}  e^{-i2\pi \frac{jm} N} \sum_{n=0}^{N-1} e^{-i2\pi \frac{kn}{N} }  A_{c,m,n}.    
\end{equation}
Since DFT depends on direction, it is not invariant to transposition, i.e., $\mathcal{F}(A_c) \neq \mathcal{F}(A_c^T)$.

Direction dependence makes DFT different from embeddings used in INR models, which treat all inputs symmetrically. Consequently, DFT has an unnecessarily complex structure (a matrix instead of a vector), capturing differences, such as image transposition, that are irrelevant to our application. To make the DFT more suitable for our purposes, we reduce it to a \mbox{\textbf{spectrum}} vector $\spectrum(A) = (\spectrum(A, 1), \dots, \spectrum(A, N-1))$ which removes direction dependence of DFT ($S(A) = S(A^T)$), for examples, see \Cref{fig:fresh_workflow_and_wasserstein}. We achieve this by summing together elements of DFT that represent the same frequency, but not direction, meaning the elements where DFT indices sum to the same number. This is well illustrated by transposing an image, as $\mathcal{F}_{i,j}(A^T_c)$ is the same as $ \mathcal{F}_{j,i}(A_c)$. Specifically, the elements of the spectrum vector are created by summing along DFT diagonals: \begin{equation}
    \label{eq:spectrum_full}
    \spectrum(A, d)=\sum_{c \in \{0,\dots, C-1\}} \sum_{\substack{(i,j) \in \{0,\dots, N-1\}\\i+j=d}} |\mathcal{F}_{i, j}(A_c)|,\\
\end{equation}
We additionally denote the first $n$ entries of $\spectrum(A)$ as:
\begin{equation}
    \label{eq:spectrum_cropped}
    \spectrum_n(A) = ( \spectrum(A, 1), \dots, \spectrum(A, n)).\\
    \end{equation}
 We omit the term $S(A, 0)$ corresponding to a constant signal, as changes to this component can be fully captured by the bias of the output layer of a neural network. As such, $\spectrum(A, 0)$ is easy to model and is not affected by changes to the embedding. 

The term \textit{spectrum} is sometimes used to describe the 2-dimensional DFT of an image. However, to the best of our knowledge, it has not been widely applied in the context of INRs, and no prior work has introduced the spectrum as a vector.

\textbf{Wasserstein distance} is a distance function between probability distributions. 
For distributions P and Q over $\real$, and a cost function $c(x,y) = |x - y|$, it is defined as:
\begin{equation}
\mathcal{W}(P, Q) = \inf_{\pi \in \Gamma(P, Q)} \int_{\real \times \real} c(x, y) \, d\pi(x,y)\,,    \end{equation}
where $\Gamma(P, Q)$ is the set of all joint distributions on $\real \times \real$ with marginals $P$ and $Q$.
Although complex to estimate in multiple dimensions, in our 1-dimensional, discrete case the Wasserstein distance is the L1 norm of the difference between cumulative distribution functions:  $ || \text{CDF}(Q) - \text{CDF}(P)||_1$\citep{panaretos2019statistical}.

\begin{table}[ht]
  
  \caption{Average PSNR on image representation tasks. \our{} outperforms or matches the baseline performance without introducing significant computational costs (see \Cref{tab:image_fitting_times}). Results for \wire{} are provided for reference only, as it is incompatible with \our{} (see \cref{sec:preliminaries}). The best results in each section are bolded.  Results are averaged over 3 seeds.} 
  
  \label{tab:img_fitting_dataset_results}
  \centering

    \begin{tabular}{@{\hspace{2.5mm}}l@{\hspace{2.5mm}}c@{\hspace{2.5mm}}c@{\hspace{2.5mm}}c@{\hspace{2.5mm}}c@{\hspace{2.5mm}}c@{\hspace{2.5mm}}c@{\hspace{2.5mm}}}
    \toprule
 \small PSNR $\uparrow$    & \small Average & \small  \chest{} & \small  \ffhqcropped{} & \small  \ffhq{} & \small  \small  \kodak{} & \small \art{} \\

\midrule
    \small \siren{} & \small $33.85$ $\mathsmaller{\pm  0.01}$ & \small $37.35$ $\mathsmaller{\pm {>} 0.01}$ & \small $37.54$ $\mathsmaller{\pm  0.04}$ & \small $34.32$ $\mathsmaller{\pm  0.01}$ & \small $31.60$ $\mathsmaller{\pm  0.03}$ & \small $28.45$ $\mathsmaller{\pm  0.01}$ \\
    \small +\our{} & \small $\mathbf{34.62}\mathsmaller{\pm  0.01}$ & \small $\mathbf{37.99}\mathsmaller{\pm  0.01}$ & \small $\mathbf{39.11}\mathsmaller{\pm  0.01}$ & \small $\mathbf{35.40}\mathsmaller{\pm  0.01}$ & \small $\mathbf{31.78}\mathsmaller{\pm  0.02}$ & \small $\mathbf{28.80}\mathsmaller{\pm  0.01}$ \\
    \midrule
    \small \ffeatures{} & \small $32.12$ $\mathsmaller{\pm  0.01}$ & \small $36.96$ $\mathsmaller{\pm  0.03}$ & \small $35.01$ $\mathsmaller{\pm  0.04}$ & \small $32.65$ $\mathsmaller{\pm  0.01}$ & \small $28.84$ $\mathsmaller{\pm  0.04}$ & \small $27.15$ $\mathsmaller{\pm  0.01}$ \\
    \small +\our{} & \small $\mathbf{33.45}\mathsmaller{\pm  0.02}$ & \small $\mathbf{37.77}\mathsmaller{\pm  0.04}$ & \small $\mathbf{36.81}\mathsmaller{\pm  0.06}$ & \small $\mathbf{34.62}\mathsmaller{\pm  0.01}$ & \small $\mathbf{30.06}\mathsmaller{\pm  0.01}$ & \small $\mathbf{28.01}\mathsmaller{\pm  0.02}$ \\
    \midrule
    \small \finer{} & \small $\mathbf{35.11}\mathsmaller{\pm {>} 0.01}$ & \small $\mathbf{38.63}\mathsmaller{\pm  0.02}$ & \small $\mathbf{40.45}\mathsmaller{\pm  0.01}$ & \small $\mathbf{36.48}\mathsmaller{\pm  0.02}$ & \small $\mathbf{31.40}\mathsmaller{\pm  0.02}$ & \small $\mathbf{28.57}\mathsmaller{\pm  0.01}$ \\
    \small +\our{} & \small $35.03$ $\mathsmaller{\pm  0.01}$ & \small $38.51$ $\mathsmaller{\pm  0.04}$ & \small $40.31$ $\mathsmaller{\pm  0.07}$ & \small $\mathbf{36.48}\mathsmaller{\pm  0.01}$ & \small $31.31$ $\mathsmaller{\pm  0.03}$ & \small $28.54$ $\mathsmaller{\pm  0.01}$ \\
    \midrule
    \small \finer{}$_{k=0}$ & \small $34.81$ $\mathsmaller{\pm  0.01}$ & \small $\mathbf{38.44}\mathsmaller{\pm  0.01}$ & \small $39.91$ $\mathsmaller{\pm  0.02}$ & \small $35.96$ $\mathsmaller{\pm  0.01}$ & \small $\mathbf{31.43}\mathsmaller{\pm  0.02}$ & \small $28.31$ $\mathsmaller{\pm  0.01}$ \\
    \small +\our{} & \small $\mathbf{34.88}\mathsmaller{\pm  0.03}$ & \small $38.43$ $\mathsmaller{\pm  0.03}$ & \small $\mathbf{40.12}\mathsmaller{\pm  0.06}$ & \small $\mathbf{36.28}\mathsmaller{\pm  0.02}$ & \small $31.16$ $\mathsmaller{\pm  0.01}$ & \small $\mathbf{28.44}\mathsmaller{\pm  0.05}$ \\
    \midrule
    \small \wire{} & \small $33.54$ $\mathsmaller{\pm  0.02}$ & \small $37.96$ $\mathsmaller{\pm  0.02}$ & \small $38.13$ $\mathsmaller{\pm  0.04}$ & \small $35.04$ $\mathsmaller{\pm  0.01}$ & \small $28.95$ $\mathsmaller{\pm  0.06}$ & \small $27.62$ $\mathsmaller{\pm  0.01}$ \\

\bottomrule
\end{tabular}    
    
\end{table}

\subsection{\our{}}

\our{} performs a parameter sweep in which the Discrete Fourier Transform and the Wasserstein distance are used to approximate model performance, instead of the costly model training required for grid search.

\textbf{Method description.} Our goal is to select an embedding configuration $\theta_i$ from a set of $M$ configurations $\{\theta_i \}_{i \in \{1, \dots, M\}}$ that would maximize performance when fitting a target image $Y \in \real^{C \times N \times N}$. We propose to use the configuration $\theta_i$ where the associated output of the model at initialization, $\hat{Y}^i_{\text{init}}$, has a similar frequency distribution to the target image, in other words $\spectrum_n(Y) \approx \spectrum_n(\hat{Y}^i_{\text{init}})$. 

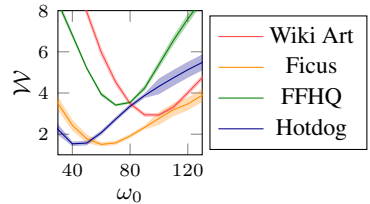
\begin{wrapfigure}{r}{0.38\linewidth}
    \centering
    \vspace{1mm}
    \begin{tikzpicture}
    \begin{axis}[
        xlabel={$\omega_0$},
        ylabel={$\mathcal{W}$},
        width=3.5cm,
        height=3.5cm,
        xmin=30,
        xmax=130,
        ymax=8,
        ymin=1,
        legend pos=north east,
        ylabel style={font=\footnotesize, yshift=-7.5mm},
        xlabel style={font=\footnotesize, yshift=2.0mm},
        legend style={
                    at={(1.05, 0.5)},
                    anchor=west,
                    font=\footnotesize,
        },
        tick label style={font=\scriptsize},
        xtick={40,80,120},
    ]
        \pgfplotstableread[col sep=comma]{example_spectra/wasserstein_nerf_siren_crop_32.csv}\datatable

        \addplot[color3] table[x=omega, y=initial_wasserstein_cropped_64, col sep=comma] {example_spectra/wasserstein_wiki_art.csv};
        \addlegendentry{\art{}}

        \addplot[sirencolor] table[x=omega, y=FICUS] {\datatable};
        \addlegendentry{Ficus}

        \addplot[freshcolor] table[x=omega, y=initial_wasserstein_cropped_64, col sep=comma] {example_spectra/wasserstein_ffhq_wild.csv};
        \addlegendentry{FFHQ}

        \addplot[baselinecolor] table[x=omega, y=HOTDOG] {\datatable};
        \addlegendentry{Hotdog}

        \addplot[mark=none,opacity=0, name path=ficus_top]
            table[x=omega, y expr=\thisrow{FICUS} + 1.96*\thisrow{FICUS SE}, col sep=comma] {example_spectra/wasserstein_nerf_siren_crop_32.csv};
        
        \addplot[mark=none,opacity=0, name path=ficus_bot]
            table[x=omega, y expr=\thisrow{FICUS} - 1.96*\thisrow{FICUS SE}, col sep=comma] {example_spectra/wasserstein_nerf_siren_crop_32.csv};
        \addplot[sirencolor, fill=sirencolor, opacity=0.3]
            fill between[of=ficus_bot and ficus_top];
            
        \addplot[mark=none,opacity=0, name path=hotdog_top]
            table[x=omega, y expr=\thisrow{HOTDOG} + 1.96*\thisrow{HOTDOG SE}, col sep=comma] {example_spectra/wasserstein_nerf_siren_crop_32.csv};
        \addplot[mark=none,opacity=0, name path=hotdog_bot]
            table[x=omega, y expr=\thisrow{HOTDOG} - 1.96*\thisrow{HOTDOG SE}, col sep=comma] {example_spectra/wasserstein_nerf_siren_crop_32.csv};
        \addplot[baselinecolor, fill=baselinecolor, opacity=0.3]
            fill between[of=hotdog_bot and hotdog_top];

        \addplot[mark=none,opacity=0, name path=t1]
            table[x=omega, y expr=\thisrow{initial_wasserstein_cropped_64} + 1.96*\thisrow{initial_wasserstein_cropped_64_se}, col sep=comma] {example_spectra/wasserstein_ffhq_wild.csv};
        \addplot[mark=none,opacity=0, name path=t2]
            table[x=omega, y expr=\thisrow{initial_wasserstein_cropped_64} - 1.96*\thisrow{initial_wasserstein_cropped_64_se}, col sep=comma] {example_spectra/wasserstein_ffhq_wild.csv};
        \addplot[freshcolor, fill=freshcolor, opacity=0.3]
            fill between[of=t1 and t2];

        \addplot[mark=none,opacity=0, name path=t1]
            table[x=omega, y expr=\thisrow{initial_wasserstein_cropped_64} + 1.96*\thisrow{initial_wasserstein_cropped_64_se}, col sep=comma] {example_spectra/wasserstein_wiki_art.csv};
        \addplot[mark=none,opacity=0, name path=t2]
            table[x=omega, y expr=\thisrow{initial_wasserstein_cropped_64} - 1.96*\thisrow{initial_wasserstein_cropped_64_se}, col sep=comma] {example_spectra/wasserstein_wiki_art.csv};
        \addplot[color3, fill=color3, opacity=0.3]
            fill between[of=t1 and t2];
        
    \end{axis}
\end{tikzpicture}
    \vspace{-1mm}
    \caption{Wasserstein distance for selected image and NeRF datasets across different \siren{} configurations. It follows a smooth trend with a distinct global minimum, indicating stable and predictable behavior. Shaded area represents the 95\% confidence interval.}
    \label{fig:wasserstein_distances}
    \vspace{-10mm}
\end{wrapfigure}
We note that the spectrum $\spectrum_n(A)$ is absolutely homogeneous ($\spectrum_n(\alpha A)=|\alpha|\spectrum_n(A)$), which implies that scaling does not affect the relative presence of different frequencies. Additionally, it is equivalent to scaling the original signal, a common data pre-processing step. As such, the spectrum can be interpreted as a probability distribution, making the Wasserstein distance a natural choice for a similarity measure. This requires only that we use the normalized spectrum, defined as $\Tilde{\spectrum_n}(A) = \frac{\spectrum_n(A)}{\|\spectrum_n(A)\|_1}$.
With this, we define the \our{} configuration $\theta_j$ as the one that minimizes the Wasserstein distance between the target signal and the model, meaning 
\begin{align}
     j &= {\argmin_{i}} \,  \mathcal{W}(\mathcal{\Tilde{S}}_n(Y), \mathcal{\Tilde{S}}_n(\hat{Y}^i_{\text{init}})) \\
     &= {\argmin_{i}} ||\text{CDF}({\mathcal{\Tilde{S}}_n(Y)}) - \text{CDF}({\Tilde{S}_n(\hat{Y}^i_{\text{init}})}) ||_1. 
\end{align}
In settings where multiple target images are available (video approximation and NeRF), we select one at random to calculate the Wasserstein distance. We visualize the entire selection process for images in \Cref{fig:fresh_workflow_and_wasserstein}, and provide a full algorithm of \our{} in the Appendix (see \cref{alg:fresh}).

\begin{figure}[t]
    \centering
    \includegraphics[width=\linewidth]{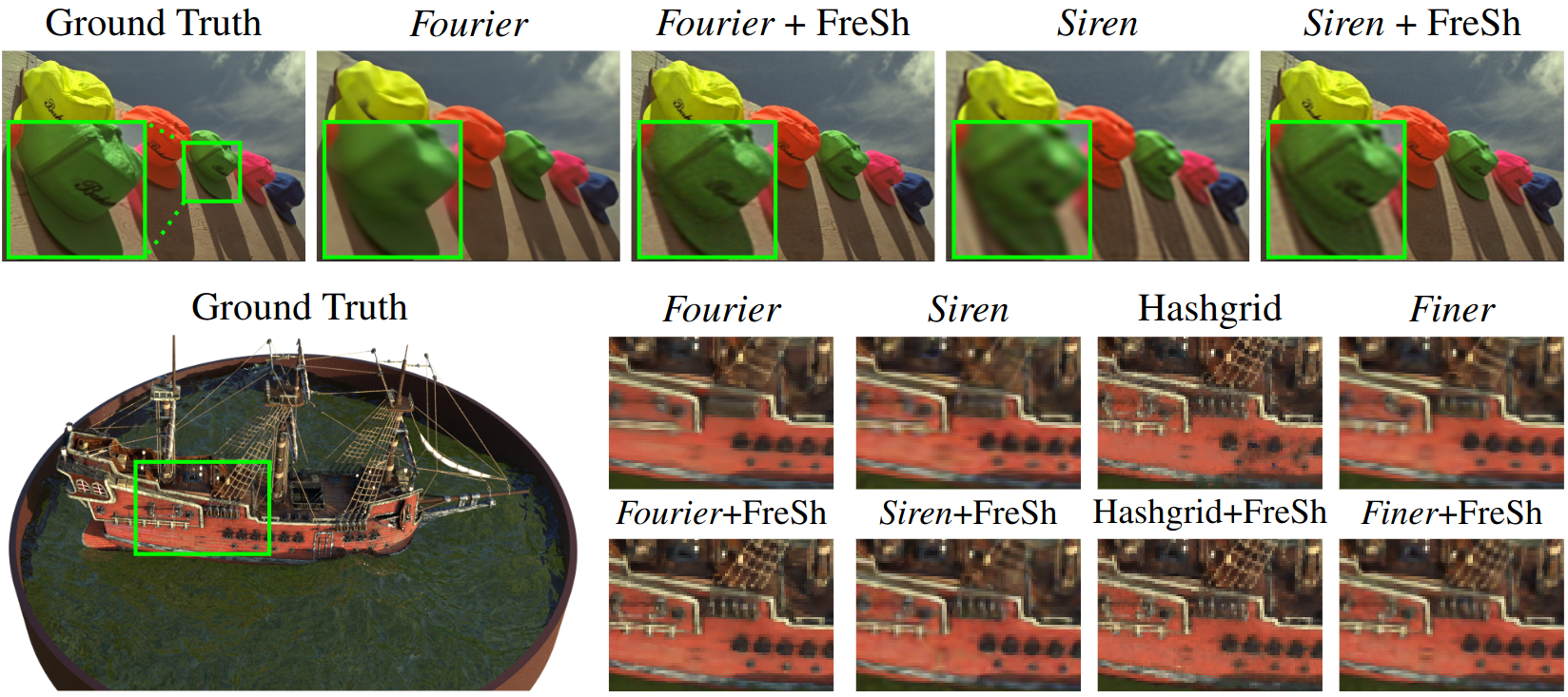}
    \caption{Example model outputs for image modeling (top) and NeRF (bottom). \our{} representations are better at modeling high-frequency details such as text or ropes. For additional examples, see \cref{app:experiments}.}
    \label{fig:image_comparision}
\end{figure}

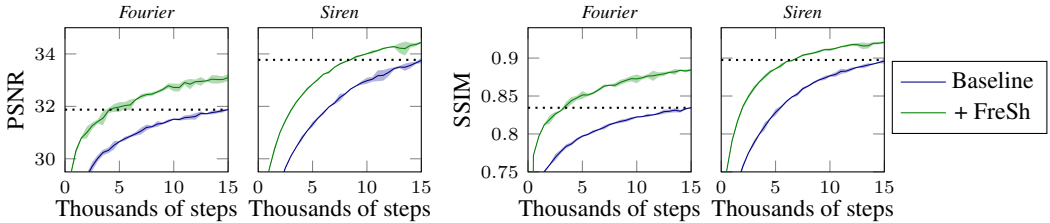
\begin{figure}
    \centering
        \begin{tikzpicture}
    \begin{groupplot}[
        group style={
            group size=5 by 1,
            horizontal sep=0.4cm,
            vertical sep=0.1cm,
        },
        width=37.5mm,
        height=35mm,
        xlabel near ticks, 
        ylabel near ticks,
        legend style={at={(2.0, 0.77)}, font=\footnotesize},
        tick label style={font=\scriptsize},
        ylabel style={font=\footnotesize, yshift=-1.0mm},
        xlabel style={font=\footnotesize, yshift=1.5mm},
        xmin=0,
        xmax=15,
    ]
    \nextgroupplot[ylabel=PSNR,  xlabel=Thousands of steps, scaled x ticks=false, ymin=29.5, ymax=35]
            \addplot[mark=none,baselinecolor] table[x expr=\thisrowno{0}*0.001, y index=1, col sep=comma] {data/plots_seeded/relu/baseline/psnr_test_results_seeded.csv};
            \addplot[mark=none,freshcolor] table[x expr=\thisrowno{0}*0.001, y index=1, col sep=comma] {data/plots_seeded/relu/fresh/psnr_test_results_seeded.csv}; 
            
            \addplot[mark=none,opacity=0, name path=path_top1] 
                table[x expr=\thisrowno{0}*0.001, y expr=\thisrowno{1}+2.58*\thisrowno{2}, col sep=comma] {data/plots_seeded/relu/baseline/psnr_test_results_seeded.csv};
            \addplot[mark=none,opacity=0, name path=path_bot1] 
                table[x expr=\thisrowno{0}*0.001, y expr=\thisrowno{1}-2.58*\thisrowno{2}, col sep=comma] {data/plots_seeded/relu/baseline/psnr_test_results_seeded.csv};
            \addplot[baselinecolor, fill=baselinecolor, opacity=0.3]
                fill between[
                    of=path_bot1 and path_top1
                ];

            \addplot[mark=none,opacity=0, name path=path_top2]
                table[x expr=\thisrowno{0}*0.001, y expr=\thisrowno{1}+2.58*\thisrowno{2}, col sep=comma] {data/plots_seeded/relu/fresh/psnr_test_results_seeded.csv}; 
            \addplot[mark=none,opacity=0, name path=path_bot2] 
                table[x expr=\thisrowno{0}*0.001, y expr=\thisrowno{1}-2.58*\thisrowno{2}, col sep=comma] {data/plots_seeded/relu/fresh/psnr_test_results_seeded.csv}; 
            \addplot[freshcolor, fill=freshcolor, opacity=0.3]
                fill between[
                    of=path_bot2 and path_top2
                ];

            \addplot[dotted, black, thick] coordinates {(0,31.874) (15,31.874)};

        \nextgroupplot[xlabel=Thousands of steps, scaled x ticks=false, ymin=29.5, ymax=35,yticklabel=\empty]
            \addplot[mark=none,baselinecolor] table[x expr=\thisrowno{0}*0.001, y index=1, col sep=comma] {data/plots_seeded/siren/baseline/psnr_test_results_seeded.csv}; 
            
            \addplot[mark=none,freshcolor] table[x expr=\thisrowno{0}*0.001, y index=1, col sep=comma] {data/plots_seeded/siren/fresh/psnr_test_results_seeded.csv};

            \addplot[mark=none,opacity=0, name path=path_top1] 
                table[x expr=\thisrowno{0}*0.001, y expr=\thisrowno{1}+2.58*\thisrowno{2}, col sep=comma] {data/plots_seeded/siren/baseline/psnr_test_results_seeded.csv}; 
            \addplot[mark=none,opacity=0, name path=path_bot1] 
                table[x expr=\thisrowno{0}*0.001, y expr=\thisrowno{1}-2.58*\thisrowno{2}, col sep=comma] {data/plots_seeded/siren/baseline/psnr_test_results_seeded.csv}; 
            
            \addplot[baselinecolor, fill=baselinecolor, opacity=0.3]
                fill between[
                    of=path_bot1 and path_top1
                ];

            \addplot[mark=none,opacity=0, name path=path_top2]
                table[x expr=\thisrowno{0}*0.001, y expr=\thisrowno{1}+2.58*\thisrowno{2}, col sep=comma] {data/plots_seeded/siren/fresh/psnr_test_results_seeded.csv};  
            \addplot[mark=none,opacity=0, name path=path_bot2] 
                table[x expr=\thisrowno{0}*0.001, y expr=\thisrowno{1}-2.58*\thisrowno{2}, col sep=comma] {data/plots_seeded/siren/fresh/psnr_test_results_seeded.csv};  
            
            \addplot[freshcolor, fill=freshcolor, opacity=0.3]
                fill between[
                    of=path_bot2 and path_top2
                ];

            \addplot[dotted, black, thick] coordinates {(0,33.774) (15,33.774)};

    \nextgroupplot[width=22mm, opacity=0]

    \nextgroupplot[ylabel=SSIM, xlabel=Thousands of steps,  scaled x ticks=false, ymin=0.75, ymax=0.94]
            \addplot[mark=none,baselinecolor] table[x expr=\thisrowno{0}*0.001, y index=1, col sep=comma] {data/plots_seeded/relu/baseline/ssim_results_seeded.csv};
            \addplot[mark=none,freshcolor] table[x expr=\thisrowno{0}*0.001, y index=1, col sep=comma] {data/plots_seeded/relu/fresh/ssim_results_seeded.csv};  
            
            \addplot[mark=none,opacity=0, name path=path_top3] 
                table[x expr=\thisrowno{0}*0.001, y expr=\thisrowno{1}+2.58*\thisrowno{2}, col sep=comma] {data/plots_seeded/relu/baseline/ssim_results_seeded.csv};
            \addplot[mark=none,opacity=0, name path=path_bot3] 
                table[x expr=\thisrowno{0}*0.001, y expr=\thisrowno{1}-2.58*\thisrowno{2}, col sep=comma] {data/plots_seeded/relu/baseline/ssim_results_seeded.csv};
            
            \addplot[baselinecolor, fill=baselinecolor, opacity=0.3]
                fill between[
                    of=path_bot3 and path_top3
                ];

            \addplot[mark=none,opacity=0, name path=path_top4]
                table[x expr=\thisrowno{0}*0.001, y expr=\thisrowno{1}+2.58*\thisrowno{2}, col sep=comma] {data/plots_seeded/relu/fresh/ssim_results_seeded.csv};
            \addplot[mark=none,opacity=0, name path=path_bot4] 
                table[x expr=\thisrowno{0}*0.001, y expr=\thisrowno{1}-2.58*\thisrowno{2}, col sep=comma] {data/plots_seeded/relu/fresh/ssim_results_seeded.csv};
            
            \addplot[freshcolor, fill=freshcolor, opacity=0.3]
                fill between[
                    of=path_bot4 and path_top4
                ];

        \addplot[dotted, black, thick] coordinates {(0,0.8345) (15,0.8345)};

        \nextgroupplot[ylabel={}, xlabel=Thousands of steps, scaled x ticks=false, ymin=0.75, ymax=0.94,yticklabel=\empty]
            \addplot[mark=none,baselinecolor] table[x expr=\thisrowno{0}*0.001, y index=1, col sep=comma] {data/plots_seeded/siren/baseline/ssim_results_seeded.csv};
            \addlegendentry{Baseline}
            
            \addplot[mark=none,freshcolor] table[x expr=\thisrowno{0}*0.001, y index=1, col sep=comma] {data/plots_seeded/siren/fresh/ssim_results_seeded.csv};
            \addlegendentry{+ \our{}}
            
            \addplot[mark=none,opacity=0, name path=path_top5] 
                table[x expr=\thisrowno{0}*0.001, y expr=\thisrowno{1}+2.58*\thisrowno{2}, col sep=comma] {data/plots_seeded/siren/baseline/ssim_results_seeded.csv};
            \addplot[mark=none,opacity=0, name path=path_bot5] 
                table[x expr=\thisrowno{0}*0.001, y expr=\thisrowno{1}-2.58*\thisrowno{2}, col sep=comma] {data/plots_seeded/siren/baseline/ssim_results_seeded.csv};
            
            \addplot[baselinecolor, fill=baselinecolor, opacity=0.3]
                fill between[
                    of=path_bot5 and path_top5
                ];

            \addplot[mark=none,opacity=0, name path=path_top6]
                table[x expr=\thisrowno{0}*0.001, y expr=\thisrowno{1}+2.58*\thisrowno{2}, col sep=comma] {data/plots_seeded/siren/fresh/ssim_results_seeded.csv};
            \addplot[mark=none,opacity=0, name path=path_bot6] 
                table[x expr=\thisrowno{0}*0.001, y expr=\thisrowno{1}-2.58*\thisrowno{2}, col sep=comma] {data/plots_seeded/siren/fresh/ssim_results_seeded.csv};
            
            \addplot[freshcolor, fill=freshcolor, opacity=0.3]
                fill between[
                    of=path_bot6 and path_top6
                ];

             \addplot[dotted, black, thick] coordinates {(0,0.8975) (15,0.8975)};

    \end{groupplot}
    \node[above] at ($(group c1r1.north)$) {\scriptsize \ffeatures{}};
    \node[above] at ($(group c2r1.north)$) {\scriptsize \siren{}};
    \node[above] at ($(group c4r1.north)$) {\scriptsize \ffeatures{}};
    \node[above] at ($(group c5r1.north)$) {\scriptsize \siren{}};
    
    \end{tikzpicture}
        \vspace{-4mm}
        \caption{Mean PSNR and SSIM values during training on 50 images (averaged over 3 seeds). \our{} improves the final performance and speeds up convergence. Dotted lines indicate the final results of the baseline model. Shaded area is the 99\% confidence interval.}
        \label{fig:image_metrics_over_time}
        \vspace{-3mm}
\end{figure}

\textbf{Measurement noise.} Due to the randomness of the image used as the target signal, $Y$, on video approximation and NeRF tasks and the randomness of the model output $\hat{Y}^i_{\text{init}}$ arising from random network weights, the measurement of the Wasserstein distance is noisy. To prevent this from affecting the selection process, we measure the Wasserstein distance 10 times and use  its mean to select the optimal configuration. This is particularly important on video approximation and NeRF tasks as there both $Y$ and $\hat{Y}^i_{\text{init}}$ are sources of noise. This results in higher measurement variance in those tasks (see \Cref{fig:wasserstein_distances}). The \our{} process could be further optimized to use less compute by decreasing the  number of measurements, especially on low-noise image representation tasks, but we do not investigate it in detail as \our{} already requires only a negligible amount of compute.

\textbf{Spectrum size.} The reason for using a cropped spectrum (\eqref{eq:spectrum_cropped}), instead of the full spectrum
(\eqref{eq:spectrum_full}), is due to the noise present in real-world signals. Synthetic signals such as $\hat{Y}_{\text{init}}$ have predominately low-frequency components (see \Cref{fig:fresh_workflow_and_wasserstein}), which makes them inherently mismatched with real-world signals in the high-frequency range. As such, using \eqref{eq:spectrum_full} (or increasing the spectrum size $n$) would shift \our{} embeddings towards higher frequencies. On the other hand, low values of $n$ can remove high-frequency (not noise) components of the target signal, leading to worse performance. We found setting $n$ to $64$ achieves good results, with additional improvements possible through adjusting $n$, which we further investigate in an ablation study in the Appendix. Although our method introduces a new hyperparameter, it is not sensitive to the target signal, making it easy to configure. Additionally, the spectrum size could be used as an implicit regularizer, as lowering it leads to lower frequencies being selected.

\section{Experiments}
\label{sec:experiments}

\begin{wraptable}{r}{0.50\textwidth}
        \centering
      \vspace{-16mm}
      \caption{Total time (h) needed to train all models (\siren{}, \ffeatures{}, \finer{}, $\text{\finer{}}_{k=0}$) for the image fitting task with grid search and \our{}, assuming that 20 hyperparameter values are tested (except \finer{} with 31 values). Time of grid search was estimated based on training time of a single configuration. \our{} is an order of magnitude faster. Measurements are averaged over 3 seeds.} 
      \label{tab:image_fitting_times}
      \vspace{-0.3mm}
      \begin{tabular}{lccc}
\toprule
\small Time (h) $\downarrow$ & \small Baseline & \small Grid Search & \small \our{} \\
\midrule
\small \siren{} & \small $29.1$ $\mathsmaller{\pm 0.3}$ & \small $584.2$ $\mathsmaller{\pm 6.6}$   & \small  ${ 30.3 }$ $\mathsmaller{\pm 0.4}$ \\
\small \ffeatures{} & \small $23.5$ $\mathsmaller{\pm 0.5}$ &  \small $470.2$ $\mathsmaller{\pm 9.4}$  & \small  ${ 24.5 }$ $\mathsmaller{\pm 0.4}$ \\
\small  \finer{} & \small $33.0 $ $\mathsmaller{\pm 0.1}$ & \small $1022.7\mathsmaller{\pm 2.2}$  & \small  ${ 36.0 }$ $\mathsmaller{\pm 0.3}$ \\
\small \finer{}$_{k=0}$ & \small $31.5 $ $\mathsmaller{\pm 0.1}$ & \small $629.6$ $\mathsmaller{\pm 2.5}$  & \small ${ 34.6 }$ $\mathsmaller{\pm 0.3}$\\
\bottomrule
\end{tabular}\\
      
\end{wraptable}

\our{} increases results quality and learning speed by tailoring the embedding to the target signal, which improves modeling across all frequencies (see \Cref{fig:img_residuals}) without enormous computational costs of conventional grid searches (see \Cref{tab:image_fitting_times}). We show performance improvements on signal representation tasks and on an inverse problem in the form of estimating radiance fields \citep{Nerf}. In all experiments, we use \our{} to select one embedding configuration from $\sigma \in \{1,2,\dots,20\}$, $\omega_0 \in \{10,20,\dots,200\}$, $\omega \in \{10,20,\dots,200\}$ and $k \in \{0.0, 0.1, \dots, 3.0\}$. Unless stated otherwise, we set the spectrum size hyperparameter, $n$, to $64$. All experiments are implemented in PyTorch \citep{paszke2019pytorch} and use the Adam optimizer \citep{kingma2014adam}.

Calculating both the Fourier transform and Wasserstein distance is computationally cheap relative to the cost of training. 
This is a crucial advantage of our method over a trail and error approach, as it makes it feasible to test multiple embedding configurations. We measured the highest relative cost of running \our{} on a high-resolution \ffhq{} image, where the time required by \our{} reached 50 seconds per tested configuration, which is equivalent to 1\% of the training time.

\subsection{Signal representation}

We evaluate on image and video overfitting using the first 10 images from FFHQ \citep{FFHQDataset} (both the "in the wild" and cropped images at a resolution of 1024x1024), \art{} \citep{WikiArtDataset}, \chest{} \citep{ChestXRayDataset}, and \kodak{} \citep{KodakDataset} datasets. For videos, we use the "bikes" and "cat" videos from \citep{SIREN}. Image heights and widths for the image fitting task range from 380 to 6720. In video overfitting, we use the current state-of-the-art solution in the form of ResFields \citep{mihajlovic2023resfields}, which improves the capacity of the MLP by making the weights time-dependent.

We report the average PSNR across all image datasets after full training (15k steps) in \Cref{tab:img_fitting_dataset_results}, example outputs in \Cref{fig:image_comparision} and training curves in \Cref{fig:image_metrics_over_time}.  \our{} consistently achieves results that are either better than or comparable to the respective baselines
, achieving similar performance about 2 times faster in the case of \siren{} and about 4 times faster in the case of Fourier features, while being an order of magnitude less costly to execute than conventional grid search (see \Cref{tab:image_fitting_times}). 
Modifying the hyperparameters of \finer{} did not lead to any noticeable improvements. Nonetheless, \our{} performs comparably well, being especially helpful when the bias is eliminated ($k=0$). 
\begin{wrapfigure}{r}{0.43\linewidth}
    
    \centering
    \begin{tikzpicture}
    \begin{axis}[
        xlabel={Frequency},
        ylabel={Relative error},
        width=5.5cm,
        height=3.55cm,
        xmin=1,
        xmax=512,
        ymax=1.5,
        ymin=0.4,
        mark size=0.3mm,
        legend pos=north east,
        ylabel style={font=\footnotesize, yshift=-4.5mm},
        xlabel style={font=\footnotesize, yshift=2mm},
        xtick={150,300,450},
        legend style={font=\scriptsize,  at={(1.0, 1.0)}},
        tick label style={font=\footnotesize},
    ]
        \addplot[freshcolor] table[x=x, y=mean_residuals, col sep=comma] {data/residuals_siren.csv};
        \addlegendentry{\siren{} + \our{}}

        \addplot[baselinecolor] table[x=x, y=mean_residuals, col sep=comma] {data/residuals_relu.csv};
        \addlegendentry{\ffeatures{} + \our{}}

        \addplot[mark=none,opacity=0, name path=siren_top]
            table[x=x, y expr=\thisrow{mean_residuals} + 1.96*\thisrow{se_residuals}, col sep=comma] {data/residuals_siren.csv};
        \addplot[mark=none,opacity=0, name path=siren_bot]
            table[x=x, y expr=\thisrow{mean_residuals} - 1.96*\thisrow{se_residuals}, col sep=comma] {data/residuals_siren.csv};
        \addplot[freshcolor, fill=freshcolor, opacity=0.3]
            fill between[of=siren_bot and siren_top];

        \addplot[mark=none,opacity=0, name path=relu_top]
            table[x=x, y expr=\thisrow{mean_residuals} + 1.96*\thisrow{se_residuals}, col sep=comma]  {data/residuals_relu.csv};
        \addplot[mark=none,opacity=0, name path=relu_bot]
            table[x=x, y expr=\thisrow{mean_residuals} - 1.96*\thisrow{se_residuals}, col sep=comma] {data/residuals_relu.csv};
        \addplot[baselinecolor, fill=baselinecolor, opacity=0.3]
            fill between[of=relu_bot and relu_top];

        \addplot[dotted] coordinates {(1, 1) (512, 1)};

    \end{axis}
\end{tikzpicture}
    \vspace{-1.8mm}
    \caption{
    The mean spectrum of residuals for the image representation task expressed as a percentage of the baseline configuration performance, as measured on \kodak{} images. The dotted line indicates the baseline performance. \our{} improves modeling across all frequencies, lowering the error by about 30\%. The shaded area represents the 95\% confidence interval.}
    \label{fig:img_residuals}
    
\end{wrapfigure}
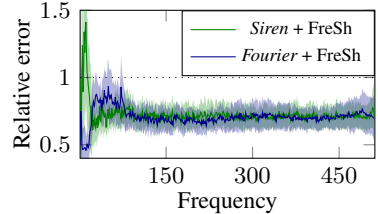
In this situation, \our{} successfully reproduces the performance of the original model with bias, raising questions about the need for both parameters. The advantage of \our{} is especially visible in terms of perceived quality, as measured by SSIM. These improvements are achieved through better approximation of the target signal across all frequency components (see \Cref{fig:img_residuals}) thanks to the increased frequency magnitudes in the embedding layer. Similarly to optimal configurations from \Cref{tab:sweep_vs_default_and_time}, configurations selected by \our{} are highly varied, highlighting the suboptimality of any constant configuration (see \Cref{app:experiments}).

To show the versatility of \our{}, we successfully apply it to videos, which contain temporal changes not directly measured by our method. We measure PSNR and SSIM and report the results in \Cref{tab:video_fitting}, with a full table available in the Appendix. We found that to improve the model's frequency modeling, we needed to remove time from the input coordinates, making it an indirect input only through the weight modifications of ResFields. This suggests that commonly used embedding strategies are ineffective for signals with qualitatively different directions that could be characterized by different frequency distributions.

\subsection{Neural Radiance Fields}

\begin{wraptable}{r}{0.43\linewidth}

  \caption{Mean PSNR on video representation. \our{} outperforms baseline embedding configurations and NeRF-like embeddings. Each configuration was tested with and without time as an input coordinate. The model benefits from embeddings reconfigured with \our{} only when time is not an input, indicating that different frequency magnitudes are required for spatial and temporal directions. Results for NeRF are provided for reference only, as it is incompatible with \our{} (see \cref{sec:preliminaries}). Results are averaged over 3 seeds.}
  \label{tab:video_fitting}
  
  \centering
        \begin{tabular}{@{\hspace{2mm}}l@{\hspace{2mm}}c@{\hspace{2mm}}c@{\hspace{2mm}}c@{\hspace{2mm}}}
        \toprule
        ~& \small Time & \small Cat & \small Bikes \\ 
        ~& \small Input  & \small PSNR $\uparrow$   & \small PSNR $\uparrow$  \\ 
        \midrule
        \small \siren{} & \small \cmark & \small $38.72$ $\mathsmaller{\pm 0.05}$  & \small $41.12$ $\mathsmaller{\pm 0.03}$  \\
        \small +\our{} & \small \cmark & \small $36.84$ $\mathsmaller{\pm 0.03}$ & \small $40.28$ $\mathsmaller{\pm 0.01}$ \\ 
        \small \siren{} & \small \xmark & \small $39.84$ $\mathsmaller{\pm 0.03}$  & \small $40.39$ $\mathsmaller{\pm 0.02}$  \\
        
        \small +\our{} & \small \xmark & \small $\mathbf{40.61 } \mathsmaller{\pm 0.04}$   & \small $\mathbf{41.62 } \mathsmaller{\pm 0.01}$   \\ 
        \hline
        \small \ffeatures{} & \small \cmark & \small $38.14$ $\mathsmaller{\pm 0.07}$ & \small $40.92$ $\mathsmaller{\pm 0.07}$  \\  
        \small +\our{}  &  \small \cmark    & \small $37.38$ $\mathsmaller{\pm 0.04}$ & \small $40.29$ $\mathsmaller{\pm 0.03}$  \\
        \small \ffeatures{} & \small \xmark & \small $38.82$ $\mathsmaller{\pm 0.03}$  & \small $ 40.61$ $\mathsmaller{\pm 0.03}$  \\  
        \small +\our{}  &  \small \xmark & \small $\mathbf{39.68 } \mathsmaller{\pm 0.04}$   & \small $\mathbf{41.13 } \mathsmaller{\pm 0.02}$  \\
        \hline
        \footnotesize Pos. enc. & \small \cmark & \small $\mathbf{37.39 } \mathsmaller{\pm 0.01}$   & \small $\mathbf{39.78 } \mathsmaller{\pm 0.04}$   \\
        \footnotesize Pos. enc. & \small \xmark & \small  $37.13$ $\mathsmaller{\pm 0.04}$  & \small $ 39.57$ $\mathsmaller{\pm 0.05}$    \\
        \bottomrule
    \end{tabular}

    \vspace{-5mm}
\end{wraptable}

Neural radiance fields \citep{Nerf} are used for synthesizing novel views of a 3D object based on a limited number of measurements (images) of the object. Unlike in the image fitting task, the target signal is unknown and spectra calculations required by \our{} are performed on the available images. To render views $\hat{Y}_{\text{init}}$ from the model, we take a single sample along each ray in the middle of the scene, where we assume the volume density is maximal. We perform experiments using a torch implementation \citep{torch-ngp,tang2022compressible} of InstantNGP \citep{muller2022instant} and synthetic NeRF data. Due to the complexity of this task, we adjusted the spectrum size for each model, and use $128$ for Hashgrid, $64$ for \ffeatures{} and \finer{}, and $32$ for the spectrally-concise \siren{}.

\our{} achieves similar or better reconstruction quality than baseline configurations and positional encoding, with the exception of axis-aligned datasets (lego, materials) where positional encodings have an advantage due to their frequencies also being axis-aligned (see \Cref{tab:nerf_results}). In the case of the Hashgrid embedding, we found that the recommended range for the resolution parameter $N_{\text{max}} \in [512, 524288]$ induced frequencies that are too high, with \our{} selecting resolutions outside this range ($N_{\text{max}}\in \{64, 128, 256\}$).
This lowering of embedding frequencies leads to  higher reconstruction quality, highlighting the applicability of \our{} in various settings and detecting models with both too low and too high frequencies.

\begin{table}[h!]
    \centering
    \caption{Average PSNR achieved with NeRF, \siren{}, \ffeatures{}, \finer{} and Hashgrid embeddings on synthetic NeRF datasets. \our{} improves performance in many instances. Results for positional encoding are provided for reference only, as it is incompatible with \our{} (see \cref{sec:preliminaries}).
    }
    
    \label{tab:nerf_results}
    \small
    \begin{tabular}{@{}l@{\hspace{1mm}}c@{\hspace{1mm}}c@{\hspace{1mm}}c@{\hspace{1mm}}c@{\hspace{1mm}}c@{\hspace{1mm}}c@{\hspace{1mm}}c@{\hspace{1mm}}c@{\hspace{1mm}}c@{}}
        \toprule
        \small PSNR $\uparrow$ &  \small Average &   \small   Chair &  \small  Drums & \small Ficus & \small Hotdog & \small Lego &   \small Materials & \small Mic &  \small Ship \\
        \midrule
            \small \siren{}                 &        $\mathsmaller{ 31.21 \pm 0.09  }$ &        $\mathsmaller{ 34.20 \pm 0.11  }$ &        $\mathsmaller{ 26.03 \pm 0.13  }$ &        $\mathsmaller{ 30.25 \pm 0.06  }$ &  $\mathsmaller{ \mathbf{35.81} \pm 0.13  }$ &        $\mathsmaller{ 30.28 \pm 0.14  }$ &  $\mathsmaller{ \mathbf{29.67} \pm 0.11  }$ &        $\mathsmaller{ 34.25 \pm 0.08  }$ &        $\mathsmaller{ 28.60 \pm 0.18  }$ \\
\small +\our{}    &  $\mathsmaller{ \mathbf{31.43} \pm 0.09}$ &  $\mathsmaller{ \mathbf{34.71} \pm 0.11  }$ &  $\mathsmaller{ \mathbf{26.15} \pm 0.14  }$ &  $\mathsmaller{ \mathbf{31.03} \pm 0.06  }$ &        $\mathsmaller{ 35.70 \pm 0.13  }$ &  $\mathsmaller{ \mathbf{30.81} \pm 0.15  }$ &        $\mathsmaller{ 29.40 \pm 0.12  }$ &  $\mathsmaller{ \mathbf{34.41} \pm 0.07  }$ &  $\mathsmaller{ \mathbf{28.74} \pm 0.19  }$ \\
\midrule
\small \ffeatures{}             &        $\mathsmaller{ 32.04 \pm 0.09  }$ &        $\mathsmaller{ 34.70 \pm 0.13  }$ &        $\mathsmaller{ 26.42 \pm 0.12  }$ &        $\mathsmaller{ 31.19 \pm 0.05  }$ &        $\mathsmaller{ 35.97 \pm 0.14  }$ &        $\mathsmaller{ 31.58 \pm 0.15  }$ &  $\mathsmaller{ \mathbf{31.67} \pm 0.10  }$ &        $\mathsmaller{ 34.38 \pm 0.09  }$ &        $\mathsmaller{ 30.07 \pm 0.12  }$ \\
\small +\our{}     &  $\mathsmaller{ \mathbf{32.63} \pm 0.09  }$ &  $\mathsmaller{ \mathbf{36.32} \pm 0.11  }$ &  $\mathsmaller{ \mathbf{26.92} \pm 0.14  }$ &  $\mathsmaller{ \mathbf{32.38} \pm 0.05  }$ &  $\mathsmaller{ \mathbf{36.26} \pm 0.14  }$ &  $\mathsmaller{ \mathbf{33.41} \pm 0.17  }$ &        $\mathsmaller{ 30.99 \pm 0.12  }$ &  $\mathsmaller{ \mathbf{34.43} \pm 0.09  }$ &  $\mathsmaller{ \mathbf{30.58} \pm 0.19  }$ \\
\midrule
\small \finer{}                 &  $\mathsmaller{ \mathbf{31.64} \pm 0.09  }$ &        $\mathsmaller{ 34.64 \pm 0.11  }$ &  $\mathsmaller{ \mathbf{26.12} \pm 0.13  }$ &  $\mathsmaller{ \mathbf{31.31} \pm 0.06  }$ &        $\mathsmaller{ 35.75 \pm 0.14  }$ &        $\mathsmaller{ 31.59 \pm 0.13  }$ &  $\mathsmaller{ \mathbf{29.71} \pm 0.11  }$ &  $\mathsmaller{ \mathbf{34.35} \pm 0.08  }$ &  $\mathsmaller{ \mathbf{29.55} \pm 0.14  }$ \\
\small +\our{}         &        $\mathsmaller{ 31.62 \pm 0.09  }$ &  $\mathsmaller{ \mathbf{34.86} \pm 0.11  }$ &        $\mathsmaller{ 26.10 \pm 0.13  }$ &        $\mathsmaller{ 31.13 \pm 0.06  }$ &  $\mathsmaller{ \mathbf{35.77} \pm 0.14  }$ &  $\mathsmaller{ \mathbf{31.80} \pm 0.13  }$ &        $\mathsmaller{ 29.51 \pm 0.11  }$ &        $\mathsmaller{ 34.27 \pm 0.07  }$ &        $\mathsmaller{ 29.47 \pm 0.17  }$ \\
\midrule
\scriptsize \finer{}$_{k=0}$         &        $\mathsmaller{ 31.39 \pm 0.09  }$ &        $\mathsmaller{ 34.38 \pm 0.11  }$ &        $\mathsmaller{ 26.02 \pm 0.13  }$ &        $\mathsmaller{ 30.80 \pm 0.06  }$ &  $\mathsmaller{ \mathbf{35.72} \pm 0.14  }$ &        $\mathsmaller{ 31.08 \pm 0.13  }$ &  $\mathsmaller{ \mathbf{29.68} \pm 0.11  }$ &        $\mathsmaller{ 34.18 \pm 0.08  }$ &        $\mathsmaller{ 28.94 \pm 0.17  }$ \\
\small +\our{} &  $\mathsmaller{ \mathbf{31.62} \pm 0.09  }$ &  $\mathsmaller{ \mathbf{34.84} \pm 0.11  }$ &  $\mathsmaller{ \mathbf{26.02} \pm 0.13  }$ &  $\mathsmaller{ \mathbf{31.23} \pm 0.06  }$ &        $\mathsmaller{ 35.66 \pm 0.14  }$ &  $\mathsmaller{ \mathbf{31.66} \pm 0.13  }$ &        $\mathsmaller{ 29.47 \pm 0.11  }$ &  $\mathsmaller{ \mathbf{34.37} \pm 0.08  }$ &  $\mathsmaller{ \mathbf{29.60} \pm 0.13  }$ \\
\midrule
\scriptsize Hashgrid                 &        $\mathsmaller{ 31.09 \pm 0.09  }$ &  $\mathsmaller{ \mathbf{34.04} \pm 0.13  }$ &        $\mathsmaller{ 25.77 \pm 0.09  }$ &        $\mathsmaller{ 30.65 \pm 0.05  }$ &        $\mathsmaller{ 34.82 \pm 0.18  }$ &        $\mathsmaller{ 31.42 \pm 0.13  }$ &        $\mathsmaller{ 28.67 \pm 0.09  }$ &  $\mathsmaller{ \mathbf{34.34} \pm 0.08  }$ &  $\mathsmaller{ \mathbf{29.06} \pm 0.18  }$ \\
\small +\our{}         &  $\mathsmaller{ \mathbf{31.22} \pm 0.09  }$ &        $\mathsmaller{ 33.55 \pm 0.14  }$ &  $\mathsmaller{ \mathbf{26.17} \pm 0.11  }$ &  $\mathsmaller{ \mathbf{31.05} \pm 0.05  }$ &  $\mathsmaller{ \mathbf{35.22} \pm 0.14  }$ &  $\mathsmaller{ \mathbf{32.56} \pm 0.14  }$ &  $\mathsmaller{ \mathbf{28.92} \pm 0.09  }$ &        $\mathsmaller{ 34.09 \pm 0.07  }$ &        $\mathsmaller{ 28.71 \pm 0.20  }$ \\
\midrule
\scriptsize Pos. enc.              &        $\mathsmaller{ 32.22 \pm 0.09  }$ &        $\mathsmaller{ 35.46 \pm 0.11  }$ &        $\mathsmaller{ 26.04 \pm 0.12  }$ &        $\mathsmaller{ 30.77 \pm 0.06  }$ &        $\mathsmaller{ 35.75 \pm 0.14  }$ &        $\mathsmaller{ 33.77 \pm 0.16  }$ &        $\mathsmaller{ 31.78 \pm 0.11  }$ &        $\mathsmaller{ 35.15 \pm 0.08  }$ &        $\mathsmaller{ 29.72 \pm 0.10  }$ \\

    \bottomrule
\end{tabular}            

\end{table}

\section{Limitations}

While this work represents an initial step toward automating embedding configuration for INRs, \our{} has several limitations inherited from existing embedding methods. Notably, \our{} does not account for direction-dependent frequency magnitudes, potentially impacting performance in scenarios where directionality is important, such as video approximation. Although the spectrum size hyperparameter is easier to configure than traditional embedding parameters due to its independence from the target signal, it still can require manual tuning. Moreover, \our{} performance is inherently limited by the constraints of the embeddings it configures and cannot achieve quality better than a conventional grid search. Specific to \our{}, it is not applicable to models that utilize excessively high embedding frequencies, such as NeRF or \wire{}.

\section{Conclusion}
\label{sec:conclusions}

Hyperparameter selection is crucial for INR performance, yet there is limited research in this area. This makes evaluating new architectures costly due to the need for extensive grid searches, or unfair when suboptimal hyperparameter values are used. We address these challenges by introducing \our{}, a model-agnostic method for configuring coordinate embeddings that significantly reduces the cost of finding effective configurations compared to parameter sweeps. \our{} leverages frequency information to select the configuration that best aligns with the target signal, effectively biasing the model to fit all relevant frequencies. While \our{} is not compatible with certain models, such as \wire{}, it proves highly effective when applicable, facilitating the use of improved embeddings like Fourier features. This is particularly relevant in the context of Neural Radiance Fields (NeRF), where the adoption of this embeddings has been limited by high hyperparameter sensitivity.
Although \our{} introduces a new hyperparameter, it is not sensitive to the target signal and it requires little to none adjustment. By utilizing ResFields, we have observed that frequency magnitudes in certain tasks are significantly direction-dependent. This suggests that new embeddings may be needed to account for this dependence, and expanding \our{} to also consider directional dependencies is a promising research direction that could further enhance its effectiveness.

\subsubsection*{Acknowledgments}

The research of A. Kania was funded by the program Excellence Initiative – Research University at the Jagiellonian University in Kraków. The work of P. Spurek was supported by the National Centre of Science (Poland) Grant No. 2023/50/E/ST6/00068

\FloatBarrier

\bibliography{iclr2025_conference}
\bibliographystyle{iclr2025_conference}

\appendix
\newpage
\section{Optimization of embedding weights}
\label{app:sgd}

\begin{wrapfigure}{r}{0.44\linewidth}
    \centering
    \begin{tikzpicture}
        \begin{groupplot}[
            group style={
                group size=1 by 1,
                horizontal sep=5mm,
            },
            width=0.35\textwidth,
            height=0.35\textwidth,
            legend pos=north east,
            ylabel style={yshift=-2mm},
            xlabel style={yshift=2mm},
            ybar,
            xtick align=inside,
            ytick align=outside,
            grid=major,
            grid style=dashed,
            every axis plot/.append style={fill opacity=0.5},
            ymax=0.35,
            xticklabel style={/pgf/number format/fixed},
        ]
        
        \nextgroupplot[ylabel={Relative frequency},xlabel={$\omega_0 \|\wrow\|_2$},bar width=1mm] 
        \addplot[bar shift=0.1cm, baselinecolor, fill=baselinecolor] table [x=BinEdges, y=Count_init, col sep=comma] {data/weight_change/siren_weights_mag_dist.csv};
        \addlegendentry{Initialised}
        \addplot[bar shift=0.1cm, orange, fill=orange] table [x=BinEdges, y=Count_final, col sep=comma] {data/weight_change/siren_weights_mag_dist.csv};
        \addlegendentry{Trained}

        \addplot[bar shift=0.1cm, freshcolor, fill=freshcolor] table [x=BinEdges, y=Count_init_fresh, col sep=comma] {data/weight_change/siren_weights_mag_dist.csv};
        \addlegendentry{Optimal}

        \end{groupplot}
    \end{tikzpicture}
    \vspace{-2mm}
    \caption{Frequency magnitudes of the embedding layer of \siren{} for baseline configurations (at initialization and after training) and an optimal configuration (at initialization) during training on 5 images from the experiment in \cref{sec:motivation}. During training, magnitudes if the baseline model fail to increase to a scale comparable with the optimal configurations, highlighting that embedding frequencies are not effectively optimized by SGD. As frequency magnitudes remain largely constant during training, they must be configured as a hyperparameter.
    Magnitudes were clipped to range $[0, 80]$.
    }
    
    \label{fig:weight_change}
\end{wrapfigure}
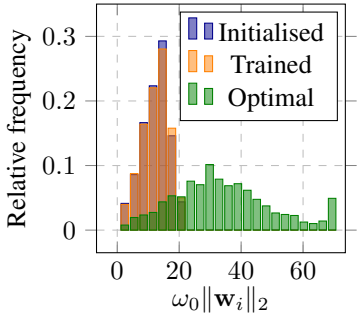

The embedding layer of an INR requires careful tuning of its hyperparameters (as discussed in \cref{sec:motivation}), even though its parameters should be optimized by SGD. To explain this phenomenon, we investigate the embedding of \siren{} and observe a pattern similar to that noted by \citet{finn2017model} about the \ffeatures{} model: SGD fails to effectively optimize the embedding layer, likely due to the periodic nature of the activations used.

We demonstrate that SGD fails to optimize the embedding layer of \siren{} by investigating the frequency magnitudes of this embedding. 
Denoting the weights of the embedding layer as $\w$ and their i-th row as $\wrow$, the magnitudes of embedding frequencies are given as $\omega_0 \|\wrow\|_2$. This follows from the form of \siren{} embedding:
\begin{equation}\label{eq:embedding_gerenic}
    \sin(\omega_0 \w \x + \mathbf{b}).
\end{equation}
We investigate how frequency magnitudes change during training on 5 images (we use the same images as in \cref{sec:motivation}) by comparing the distributions of magnitudes between models at initialization and after training (see \Cref{fig:weight_change}). We additionally provide magnitudes of embeddings configured using grid search (see \Cref{tab:sweep_vs_default_and_time}) as a reference of optimal magnitudes. During training the magnitudes change only slightly, but this increase is negligible when compared to magnitudes induced by optimal embeddings, whose size reflects the multiple-fold increase of $\omega_0$ observed in \Cref{tab:sweep_vs_default_and_time}. Since SGD fails to notably increase embedding frequencies, the magnitudes must be adjusted as a hyperparameter, significantly increasing the cost of finding optimal models.

\section{Architectures from related works}
\label{app:related}

This section covers architectures that were left out of the main text.

\label{sec:inr_architectures}

\textbf{NeRF} \citep{Nerf} maps 5D coordinates - spatial location (x, y, z) and viewing direction ($\theta$, $\varphi$) - to volume density and view-dependent emitted radiance, which are then used to render novel views of a scene. It employs positional embedding \citep{vaswani2017attention}, which is a multiresolution sequence of $L$ frequencies:
\begin{equation}\label{eq:positional_embedding}
    \gamma_P(\x) = [\sin(2^0 \x), \cos(2^1 \x), \dots, \sin(2^{L-2} \x), \cos(2^{L-1} \x)]. 
\end{equation}
$\gamma_P$ helps overcome the spectral bias of MLPs but is biased toward axis-aligned directions, which can result in performance loss depending on the rotation of the target object \citep{FourierFeatures}. This embedding works well with frequencies which are higher than the main components of the target signal (e.g., $L=16$). When training NeRF models, we embed the spatial coordinates using $\gamma_P$ and the viewing direction with the spherical harmonics basis \citep{fridovich2022plenoxels, verbin2022ref}, following the approach of \citet{muller2022instant}.

\textbf{Fourier features} \citep{FourierFeatures} is a direction-invariant alternative to \cref{eq:positional_embedding} which densely samples Fourier basis functions. Although such sampling is not feasible in realistic settings, it can be well approximated through random sampling \citep{randomFeatures}, resulting in the mapping:
\begin{equation}\label{eq:fourier_features}
    \gamma_F(\x) = [\sin(2\pi \w\x), \cos(2\pi \w\x)], 
\end{equation}
where weights are sampled from an isotropic frequency distribution, such as Gaussian $\w \sim \mathcal{N}(0, \sigma)$.  Scale of this distribution, $\sigma$, controls frequency magnitudes and it was found to be an important factor in the final performance of a model \citep{FourierFeatures}. Even though the performance of Fourier features was shown to be better than that of \cref{eq:positional_embedding}, it has not been widely adopted, possibly due to the high sensitivity of this embedding to the value of $\sigma$. We refer to this embedding as \ffeatures{}.

\textbf{\finer{}} \citep{liu2024finer} is a variation of the \siren{} model with a broader supported frequency set. It extends the frequency set of \siren{} by utilizing a variable-periodic activation, $\varphi(x)=\sin((|x|+1)x)$, in its embedding:
\begin{equation}
    \gamma_F(\x) = \varphi(\omega (\w \x + \mathbf{b})),
\end{equation}
where $\w\in \real^{m\times d}$ is the matrix of weights and $b\in \real^m$ is bias. This model controls frequencies through the scaling parameter $\omega$ and the width of bias distribution $k$ ($\mathbf{b} \sim \text{Uniform}(-k,k))$ \footnote{\citet{WIRE} denote $\omega$ as $\omega_0$. We removed the index to make hyperparameters of \finer{} and \siren{} easier to differentiate.}. The authors of \finer{} suggest using bias as the main method for increasing the model capacity to approximate high-frequency signals. However, a similar effect can also be achieved by adjusting $\omega$, which raises questions about the necessity of using bias. We consider two scenarios: one where $\omega$ is fixed at 30 and $k$ is optimized, and another where bias is removed ($k=0$) and $\omega$ is optimized. We find that bias can be removed without significantly affecting performance (see \Cref{tab:img_fitting_dataset_results}). We denote the model with no bias as $\text{\finer{}}_{k=0}$.

\textbf{Multiresolution hash encoding} (Hashgrid) divides the space into increasingly finer grids (e.g., 16), assigning each voxel to a random feature vector via a hash table \citep{muller2022instant}. The authors recommend adjusting specific hyperparameters for each dataset, particularly the size of the hash table, which impacts the memory footprint, and the resolution of the finest level, $N_{\text{max}}$, which affects the size  of details that can be easily modelled. As a baseline, we use a resolution of $2048$. At low resolutions, the number of grid vertices is of comparable size to the number of hash table entries, effectively limiting the total parameter count of low resolution models. To prevent different parameter counts from affecting the performance, we set a relatively low hash table size of 8192 entries.

\textbf{\wire{}} uses a continuous complex Gabor wavelet activation function:
\begin{equation}
    \psi(x)=e^{j\omega_{0}x}e^{-|s_{0}x|^{2}},
\end{equation}
where $\omega_{0}$ controls frequency and $s_0$ controls width of the wavelet, with typically used values of $\omega_0=20$ and $s_0=10$. This activation makes model design problematic, as it increases frequencies at every layer of the model, making frequencies of the model depth-dependent. Moreover, we note that, similarly to NeRF, \wire{} uses extremely high frequencies, which are incompatible with \our{}. As such, we report results of \wire{} only for context.

\section{Pseudocode for \our{}}
\label{app:pseudocode}

\label{sec:algorithm}

We present the pseudocode for \our{} in \Cref{alg:fresh}. The algorithm is slightly different between image and video/NeRF tasks, due to multiple images being available for the latter. This is reflected in the definition of 
 $Y_{sample}$ which consist of multiple images - one for each measurement of the Wasserstein distance. It is constructed by sampling 10 images for NeRF and video approximation tasks, while for the image  approximation task, $Y_{sample}$ consists of the same image repeated 10 times. Even though the target signal is not random for the image approximation task, it is also measured multiple times due to the randomness of the model output, $\text{model}(\theta, X)$. 

\begin{algorithm}[H]
\SetKwInput{KwInput}{Input}
\SetKwInput{KwOutput}{Output}
\SetKwInput{KwData}{Data}
\caption{\our{}}
\label{alg:fresh}
\KwInput{$\Theta$ {set of embedding configurations}, n \textgreater\ 0}
\KwData{$Y_{sample}$ a sample of images, $X$ input coordinates}
\KwOutput{$\theta_{\text{best}}$ {embedding configuration with the lowest Wasserstein distance}}

$d_{\text{best}} \gets \infty$\;
$\theta_{\text{best}} \gets \text{None}$\;

\For{$\theta \in \Theta$}{
    $distances \gets []$\;
    
    \For{$Y \text{ in } Y_{sample}$}{
        $S_{\text{target}} \gets \text{full\_spectrum}(Y)[:n]$\;
        $S_{\text{model}} \gets \text{full\_spectrum}(\text{model}(\theta, X))[:n]$\;
        $d \gets \text{wasserstein\_distance}(S_{\text{model}}, S_{\text{target}})$\;
        $distances.\text{append}(d)$\;
    }
    $d_{\text{mean}} \gets \text{mean}(distances)$\;
    \If{$d_{\text{mean}} < d_{\text{best}}$}{
        $d_{\text{best}} \gets d_{\text{mean}}$\;
        $\theta_{\text{best}} \gets \theta$\;
    }
}
\Return $\theta_{\text{best}}$\;
\end{algorithm}

\section{Additional signal representation and reconstruction results}
\label{app:experiments}

\begin{table}[ht]
  \caption{\textbf{Ablation study of spectrum size} on image approximation task. All runs in the table were conducted with a fixed seed. The optimal spectrum size depends primarily on the base architecture, with the dataset having minimal impact. The best results in each section are bolded.} 
  \label{tab:img_fitting_dataset_results_ablation}
  \centering
    \begin{tabular}{@{\hspace{2mm}}l@{\hspace{2mm}}c@{\hspace{2mm}}c@{\hspace{2mm}}c@{\hspace{2mm}}c@{\hspace{2mm}}c@{\hspace{2mm}}c@{\hspace{2mm}}c@{\hspace{3mm}}}
    \toprule
     & \small  $n$ & \small  Average & \small \chest{} & \small  \ffhqcropped{} & \small  \ffhq{} & \small  \kodak{} & \small  \art{} \\

    \midrule
    \footnotesize  \siren{} + \our{} & \footnotesize $32$ & {\footnotesize $34.28 $} & {\footnotesize ${37.49} $} & {\footnotesize ${38.13} $} & {\footnotesize ${35.00} $} & {\footnotesize $\mathbf{32.13} $} & {\footnotesize ${28.67} $} \\
    \footnotesize  \siren{} + \our{} & \footnotesize $64$ & {\footnotesize $ \mathbf{34.59}$} & {\footnotesize $ 38.00 $ } & {\footnotesize $ \mathbf{39.09}$} & {\footnotesize $ \mathbf{35.37}$} & {\footnotesize $ {31.72}$} & {\footnotesize $ \mathbf{28.78}$} \\
    \footnotesize \siren{} + \our{} & \footnotesize $128$ & {\footnotesize $ 34.15 $ } & {\footnotesize $ \mathbf{38.37}$} & {\footnotesize $ 38.66 $ } & {\footnotesize $ 34.93 $ } & {\footnotesize $ 30.41 $ } & {\footnotesize $ 28.37 $ } \\
    \midrule
    \footnotesize \ffeatures{} + \our{} & \footnotesize $32$ & {\footnotesize $33.00 $} & {\footnotesize ${37.33} $} & {\footnotesize ${36.03} $} & {\footnotesize ${34.06} $} & {\footnotesize ${29.93} $} & {\footnotesize ${27.68} $} \\
    \footnotesize \ffeatures{} + \our{} & \footnotesize $64$ & {\footnotesize $ 33.47 $ } & {\footnotesize $ 37.80 $ } & {\footnotesize $ 36.83 $ } & {\footnotesize $ 34.64$ } & {\footnotesize $ \mathbf{30.04}$} & {\footnotesize $ \mathbf{28.05}$} \\
    \footnotesize \ffeatures{} + \our{} & \footnotesize $128$ & {\footnotesize $ \mathbf{33.55}$} & {\footnotesize $ \mathbf{38.10}$} & {\footnotesize $ \mathbf{37.13}$} & {\footnotesize $ \mathbf{34.89}$} & {\footnotesize $ 29.58 $ } & {\footnotesize $ 28.03 $ } \\
    \midrule
     \footnotesize \finer{}   + \our{} & \footnotesize $32$ & {\footnotesize $34.80 $} & {\footnotesize $38.43 $} & {\footnotesize $39.87 $} & {\footnotesize $35.96 $} & {\footnotesize $\mathbf{31.44} $} & {\footnotesize $28.30 $} \\
    \footnotesize \finer{}   + \our{} & \footnotesize $64$ & {\footnotesize $ \mathbf{35.06} $ } & {\footnotesize $ 38.59 $ } & {\footnotesize $ \mathbf{40.44} $ } & {\footnotesize $ \mathbf{36.50} $ } & {\footnotesize $ {31.23} $ } & {\footnotesize $ \mathbf{28.56}$} \\
   \footnotesize \finer{}   + \our{} & \footnotesize $128$ & {\footnotesize $ 34.97 $ } & {\footnotesize $ \mathbf{38.76}$} & {\footnotesize $ 40.09 $ } & {\footnotesize $ 36.38 $ } & {\footnotesize $ 31.17 $ } & {\footnotesize $ 28.46 $ } \\

        \midrule 
    \footnotesize \finer{}$_{k=0}$ + \our{} & \footnotesize $32$ & {\footnotesize $34.44 $} & {\footnotesize $38.07 $} & {\footnotesize $39.16 $} & {\footnotesize $35.33 $} & {\footnotesize $\mathbf{31.42} $} & {\footnotesize $28.19 $} \\
    \footnotesize \finer{}$_{k=0}$ + \our{} & \footnotesize $64$ & {\footnotesize $ \mathbf{34.95}$} & {\footnotesize $ 38.50 $ } & {\footnotesize $ \mathbf{40.27}$} & {\footnotesize $ 36.32 $ } & {\footnotesize $ {31.13} $ } & {\footnotesize $ \mathbf{28.53}$} \\
    \footnotesize \finer{}$_{k=0}$ + \our{} & \footnotesize $128$ & {\footnotesize $ 34.79 $ } & {\footnotesize $ \mathbf{38.68}$} & {\footnotesize $ 40.10 $ } & {\footnotesize $ \mathbf{36.35}$} & {\footnotesize $ 30.61 $ } & {\footnotesize $ 28.24 $ } \\

\bottomrule
\end{tabular}    

\end{table}

\paragraph{Spectrum size} 
Spectrum size, $n$, controls the number of frequencies included in the spectrum (\cref{eq:spectrum_cropped}). The size of the spectrum has to be limited due to the inherent mismatch between natural and synthetic signals (as discussed in \cref{sec:method_desc}). Given that the architecture primarily determines the characteristics of the spectrum (e.g., \siren{} was designed to have a concise spectrum), the specific target signal is not expected to play a significant role in selecting the optimal spectrum size. Our ablation study in \Cref{tab:img_fitting_dataset_results_ablation} supports this, showing that the source of the dataset has minimal impact on the optimal spectrum size. Since the optimal spectrum size does not depend on the target signal, fine-tuning this value is significantly simplified, as it can be selected using a small subset of target signals. After determining the optimal spectrum size for a given architecture, no further tuning should be needed.

In general, we find that \our{} performs well when the spectrum size $n$ is set to $64$, although performance can be further improved through fine-tuning (see \Cref{tab:img_fitting_dataset_results_ablation}). 
Interestingly, \siren{}-based models gain no benefit from increasing the spectrum size beyond $64$. This is likely because \siren{} is designed with hidden layers that have minimal impact on frequencies \citep{SIREN}, leading to a concise spectrum, which is well described even with a shorter spectrum. In contrast, the \ffeatures{} model benefits from larger spectra, likely because no special considerations were made in its design to keep the spectrum concise.

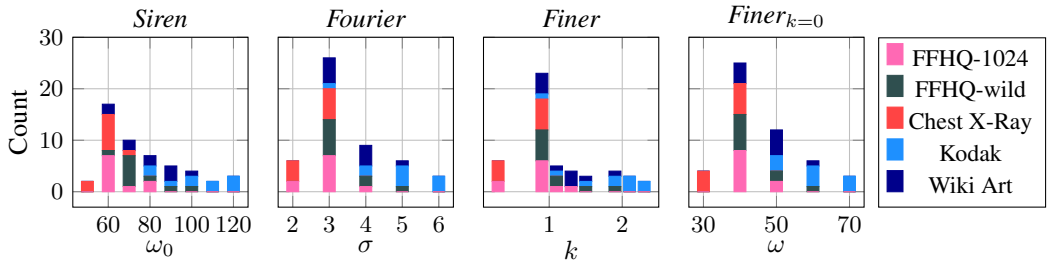
\begin{figure}
    \centering
    \vspace{-2mm}
    \begin{tikzpicture}
        \begin{groupplot}[
            group style={
                group size=4 by 1,
                horizontal sep=4mm,
            },
            width=0.28\textwidth,
            height=0.275\textwidth,
            grid=both,
            legend pos=north east,
            no marks,
            ybar stacked, 
            legend style={
                    at={(1.08,1.00)},
                    anchor=north west, 
                    font=\footnotesize,
                },
            ymax=30,
            ylabel style={yshift=-5mm},
            xlabel style={yshift=2mm},
            tick label style={font=\footnotesize},
            title style={yshift=-2.0mm},
        ]
        
        \nextgroupplot[ylabel={Count},xlabel={$\omega_0$}, title=\siren{}]
        \addplot+ [bar width=4.5pt, color4] table [x=x, y=ffhq_1024, col sep=comma] {data/configs/configs_count_siren_cropped_64.csv};

        \addplot+ [bar width=4.5pt, color1] table [x=x, y=ffhq_wild, col sep=comma] {data/configs/configs_count_siren_cropped_64.csv};

        \addplot+ [bar width=4.5pt, color3] table [x=x, y=chest_xray, col sep=comma] {data/configs/configs_count_siren_cropped_64.csv};

        \addplot+ [bar width=4.5pt, relucolor] table [x=x, y=kodak, col sep=comma] {data/configs/configs_count_siren_cropped_64.csv};

        \addplot+ [bar width=4.5pt, baselinecolor] table [x=x, y=wiki_art, col sep=comma] {data/configs/configs_count_siren_cropped_64.csv};

        \nextgroupplot[ylabel={},xlabel={$\sigma$},yticklabels={},title=\ffeatures{},xtick={2,3,4,5,6}]
        \addplot+ [bar width=4.5pt, color4] table [x=x, y=ffhq_1024, col sep=comma] {data/configs/configs_count_relu_cropped_64.csv};

        \addplot+ [bar width=4.5pt, color1] table [x=x, y=ffhq_wild, col sep=comma] {data/configs/configs_count_relu_cropped_64.csv};

        \addplot+ [bar width=4.5pt, color3] table [x=x, y=chest_xray, col sep=comma] {data/configs/configs_count_relu_cropped_64.csv};

        \addplot+ [bar width=4.5pt, relucolor] table [x=x, y=kodak, col sep=comma] {data/configs/configs_count_relu_cropped_64.csv};

        \addplot+ [bar width=4.5pt, baselinecolor] table [x=x, y=wiki_art, col sep=comma] {data/configs/configs_count_relu_cropped_64.csv};

        \nextgroupplot[ylabel={},xlabel={$k$}, title=\finer{},
        yticklabels={}]
        \addplot+ [bar width=4.5pt, color4] table [x expr=\thisrow{x}+0.1, y=ffhq_1024, col sep=comma] {data/configs/configs_count_finer_k_cropped_64.csv};

        \addplot+ [bar width=4.5pt, color1] table [x expr=\thisrow{x}+0.1, y=ffhq_wild, col sep=comma] {data/configs/configs_count_finer_k_cropped_64.csv};

        \addplot+ [bar width=4.5pt, color3] table [x expr=\thisrow{x}+0.1, y=chest_xray, col sep=comma] {data/configs/configs_count_finer_k_cropped_64.csv};

        \addplot+ [bar width=4.5pt, relucolor] table [x expr=\thisrow{x}+0.1, y=kodak, col sep=comma] {data/configs/configs_count_finer_k_cropped_64.csv};

        \addplot+ [bar width=4.5pt, baselinecolor] table [x expr=\thisrow{x}+0.1, y=wiki_art, col sep=comma] {data/configs/configs_count_finer_k_cropped_64.csv};

        \nextgroupplot[ylabel={},xlabel={$\omega$}, title=\finer{}$_{k=0}$,
        xtick={30,50,70},yticklabels={}]
        
        \addplot+ [bar width=4.5pt, color4] table [x=x, y=ffhq_1024, col sep=comma] {data/configs/configs_count_finer_cropped_64.csv};
        \addlegendentry{\ffhqcropped{}}

        \addplot+ [bar width=4.5pt, color1] table [x=x, y=ffhq_wild, col sep=comma] {data/configs/configs_count_finer_cropped_64.csv};
        \addlegendentry{\ffhq{}}

        \addplot+ [bar width=4.5pt, color3] table [x=x, y=chest_xray, col sep=comma] {data/configs/configs_count_finer_cropped_64.csv};
        \addlegendentry{\chest{}}

        \addplot+ [bar width=4.5pt, relucolor] table [x=x, y=kodak, col sep=comma] {data/configs/configs_count_finer_cropped_64.csv};
        \addlegendentry{\kodak{}}

        \addplot+ [bar width=4.5pt, baselinecolor] table [x=x, y=wiki_art, col sep=comma] {data/configs/configs_count_finer_cropped_64.csv};
        \addlegendentry{\art{}}
        
        \end{groupplot}
    \end{tikzpicture}
    \vspace{-2mm}
    \caption{Embedding configurations selected by \our{} for the image representation task. Hyperparameter values are highly varied even within datasets, highlighting the need for fine-tuning the embedding layer for each target signal.}
    \vspace{-6mm}
    \label{fig:image_fitting_configs}
\end{figure}

\paragraph{Selected hyperparameter values} We report configurations selected by \our{} on the image approximation task in \Cref{fig:image_fitting_configs}. Similarly to the optimal configurations (see \Cref{tab:sweep_vs_default_and_time}), we observe high variability in model configurations selected by \our{}. Even when narrowed to a single dataset (e.g. \kodak{}), the configurations are varied, highlighting the need for adjusting the embedding frequencies for each target signal.

\paragraph{Decreasing the default embedding frequency} In almost all experiments in \cref{sec:experiments}, models benefit from embedding configurations that induce higher frequencies than the baseline configuration. However, the Hashgrid model is an exception, where \our{} improves performance by selecting configurations that induce lower frequencies. In this section, we present another such example by testing \siren{} on a low-frequency synthetic dataset.

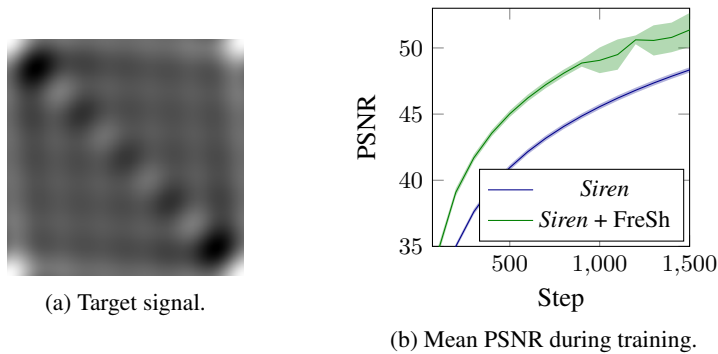
\begin{figure}[h]
    \centering
    \hspace{11mm}
    \begin{subfigure}[b]{0.45\linewidth}
        \centering
        \includegraphics[width=0.5\linewidth]{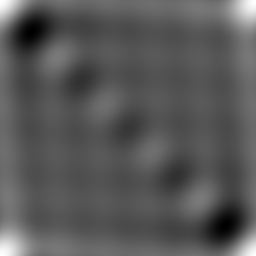}
        \caption{Target signal.}
         \vspace{5mm}
    \end{subfigure}
    \hspace{-9mm}
    \begin{subfigure}[b]{0.45\linewidth}
        \centering
        \begin{tikzpicture}
    \begin{axis}[
        xlabel={Step},
        ylabel={PSNR},
        width=5cm,
        height=4.75cm,
        xmax=1500,
        ymax=53,
        ymin=35,
        ylabel style={yshift=-3.5mm},
        xlabel style={yshift=1.0mm},
        legend style={
                    at={(0.58, 0.02)}, 
                    anchor=south, 
                    font=\footnotesize,
        },
        tick label style={font=\footnotesize},
    ]

        \addplot[baselinecolor] table[x=global_step, y=mean, col sep=comma] {data/low_freq/30_psnr_test.csv};
        \addlegendentry{\siren{}}

        \addplot[freshcolor] table[x=global_step, y=mean, col sep=comma] {data/low_freq/10_psnr_test.csv};
        \addlegendentry{\siren{} + \our{}}

        \addplot[mark=none,opacity=0, name path=ficus_top]
            table[x=global_step, y expr=\thisrow{mean} +\thisrow{196_se}, col sep=comma] {data/low_freq/30_psnr_test.csv};
        \addplot[mark=none,opacity=0, name path=ficus_bot]
            table[x=global_step, y expr=\thisrow{mean} - \thisrow{196_se}, col sep=comma] {data/low_freq/30_psnr_test.csv};
        \addplot[baselinecolor, fill=baselinecolor, opacity=0.3]
            fill between[of=ficus_bot and ficus_top];

        \addplot[mark=none,opacity=0, name path=ficus_top]
            table[x=global_step, y expr=\thisrow{mean} +\thisrow{196_se}, col sep=comma] {data/low_freq/10_psnr_test.csv};
        \addplot[mark=none,opacity=0, name path=ficus_bot]
            table[x=global_step, y expr=\thisrow{mean} - \thisrow{196_se}, col sep=comma] {data/low_freq/10_psnr_test.csv};
        \addplot[freshcolor, fill=freshcolor, opacity=0.3]
            fill between[of=ficus_bot and ficus_top];

    \end{axis}
\end{tikzpicture}
         \vspace{-1mm}
        \caption{Mean PSNR during training.}
    \end{subfigure}
    
    \caption{Training of \siren{} on an low frequency image. (a) Target image created as a sum of sinusoids of at most 5 periods. (b) PSNR values averaged over 10 seeds. \siren{} using the default configuration ($\omega_0=30$) is slower in fitting the signal than a low-frequency \siren{} configured with \our{} ($\omega_0=10$). This example demonstrates that reducing the model frequency can sometimes be advantageous, and \our{} is capable of identifying such situations.}
    \label{fig:low_frequency_training}
\end{figure}

We generate the target signal as a sum of sinusoids with up to 5 periods and train \siren{} using the same setting as described in \cref{sec:experiments}.
Since the target signal is relatively simple, we reduce the training time by a factor of 10. We found that with the default learning rate training is not stable and PSNR can decrease by us much as 40\% in 100 steps, which is why we additionally lower the learning rate by a factor of 10. 
For this dataset, \our{} selects configuration of $\omega_0=10$, resulting in frequencies three times lower than those in the baseline model. This low-frequency embedding improves the results, as shown in \Cref{fig:low_frequency_training}. This shows that \our{} improves quality by aligning frequencies and not by simply increasing them.

\begin{figure}
    \centering
    
    \begin{tikzpicture}[remember picture]
        \node at (0mm, 0mm) (baseline) {\includegraphics[width=140mm]{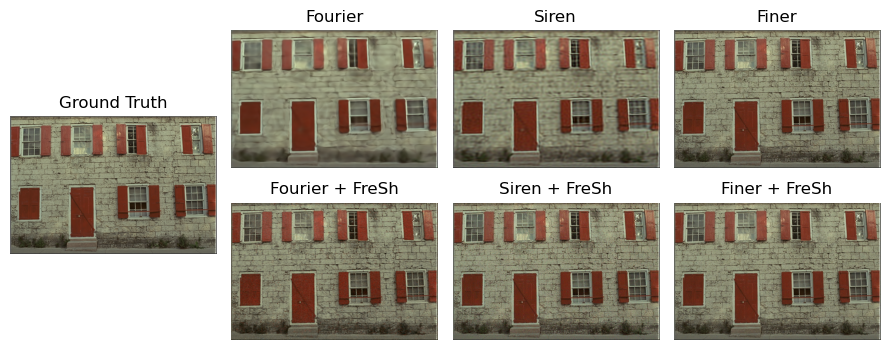}};
        \node at (0mm, -55mm) (baseline) {\includegraphics[width=140mm]{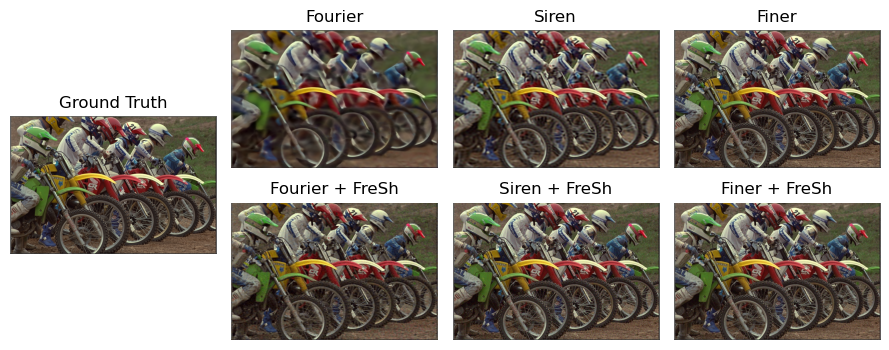}};
    \end{tikzpicture}
    
    \caption{Model comparison on \kodak{} images after 1000 iterations of training.}
    \label{fig:kodak_examples}
\end{figure}

\begin{figure}
    \centering
    
    \begin{tikzpicture}[remember picture]
        \node at (0mm, 0mm) (baseline) {\includegraphics[width=140mm]{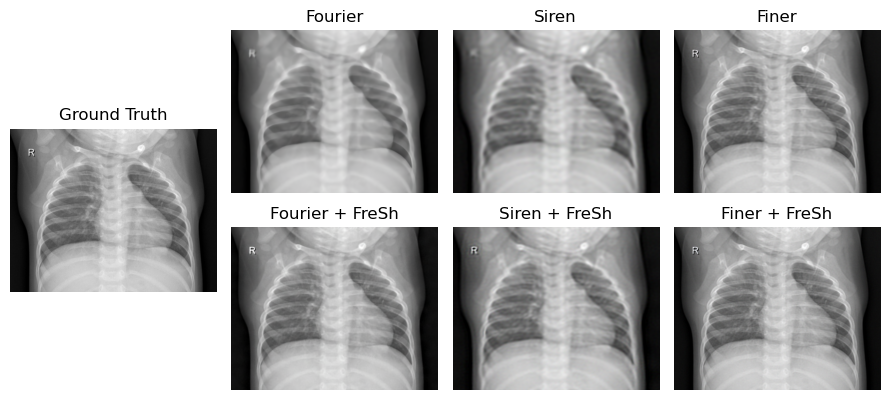}};
    \end{tikzpicture}
    
    \caption{Model comparison on a \chest{} image after 1000 iterations of training.}
    \label{fig:chest_examples}
\end{figure}

\begin{figure}
    \centering
    
    \begin{tikzpicture}[remember picture]
            \node at (0mm, 0mm) (baseline) {\includegraphics[width=140mm]{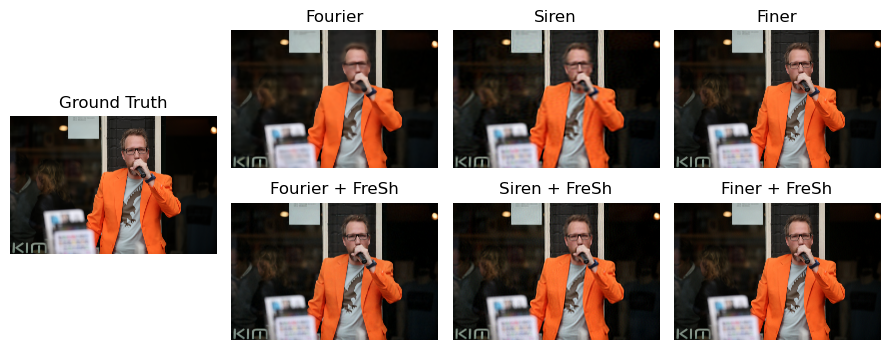}};
            \node at (0mm, -65mm) (baseline) {\includegraphics[width=140mm]{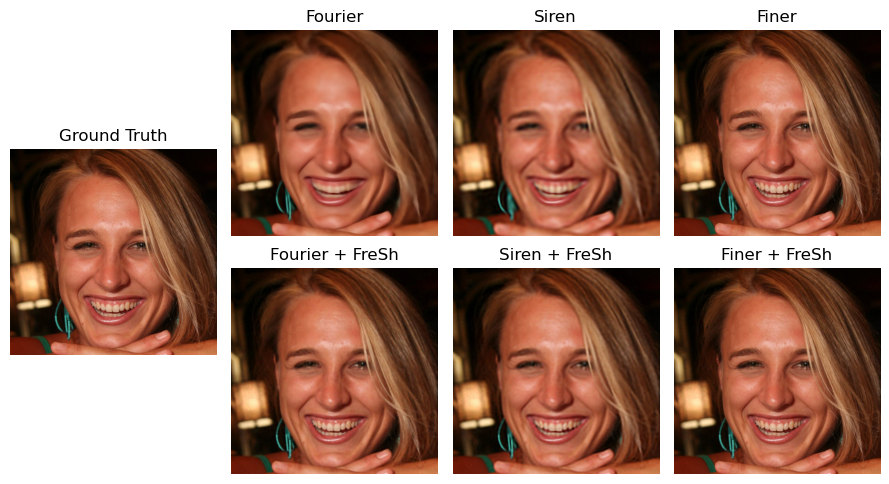}};;
    \end{tikzpicture}
    
    \caption{Model comparison on images from \ffhq{} and \ffhqcropped{} after 1000 iterations of training.}
    \label{fig:ffhq_examples}
\end{figure}

\begin{figure}
    \centering
    
    \begin{tikzpicture}[remember picture]
            \node at (0mm, 0mm) (baseline) {\includegraphics[width=140mm]{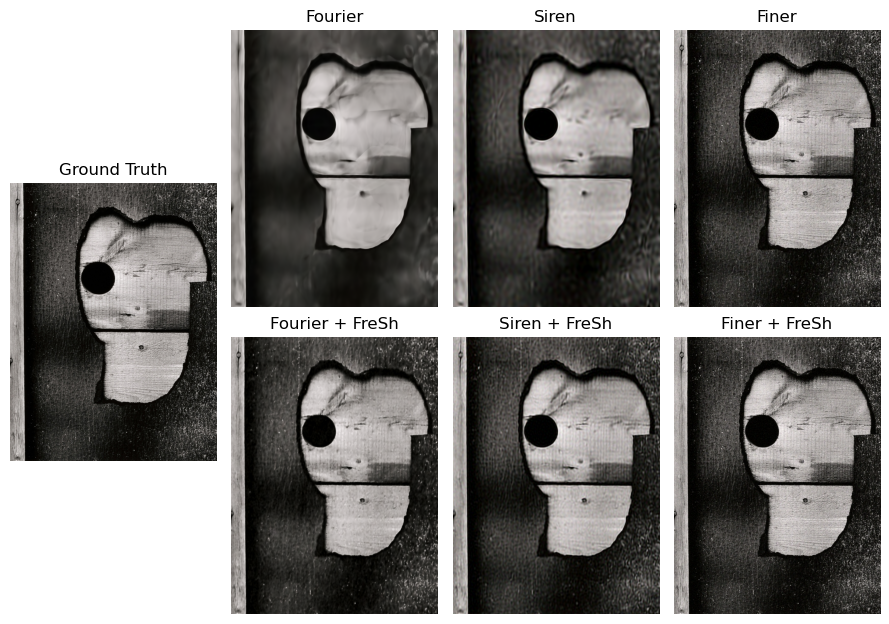}};
            \node at (0mm, -85mm) (baseline) {\includegraphics[width=140mm]{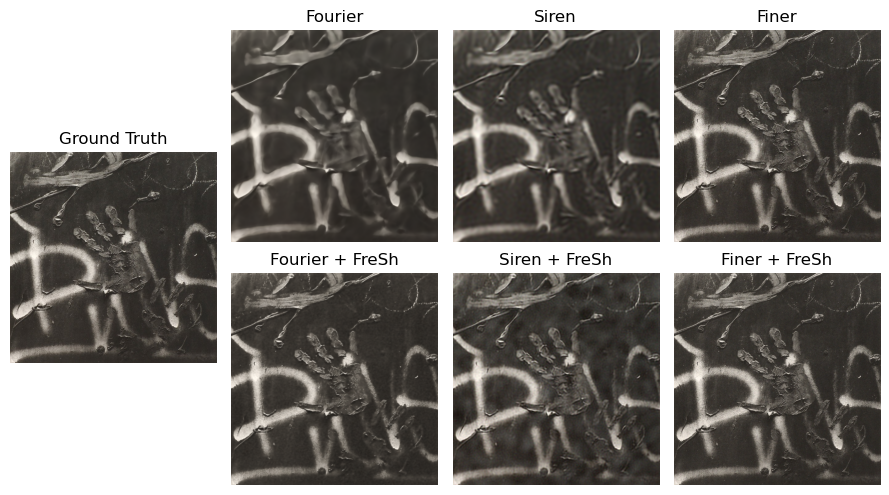}};;
    \end{tikzpicture}
    
    \caption{Model comparison on \art{} images after 1000 iterations of training.}
    \label{fig:art_examples}
\end{figure}

\paragraph{Learning speed} Re-configuring a model with \our{} increases the learning speed, with image approximations being sharp after as little as 1000 iterations. We include example outputs from all image datasets in \Cref{fig:kodak_examples,fig:ffhq_examples,fig:chest_examples,fig:art_examples}.

\paragraph{Video approximation} We report PSNR and SSIM scores on the video approximation task in \Cref{tab:video_fitting_full}. Similarly to PSNR, SSIM is highest with models configured using \our{} and without a time input.

\begin{figure}
    \includegraphics[width=\linewidth]{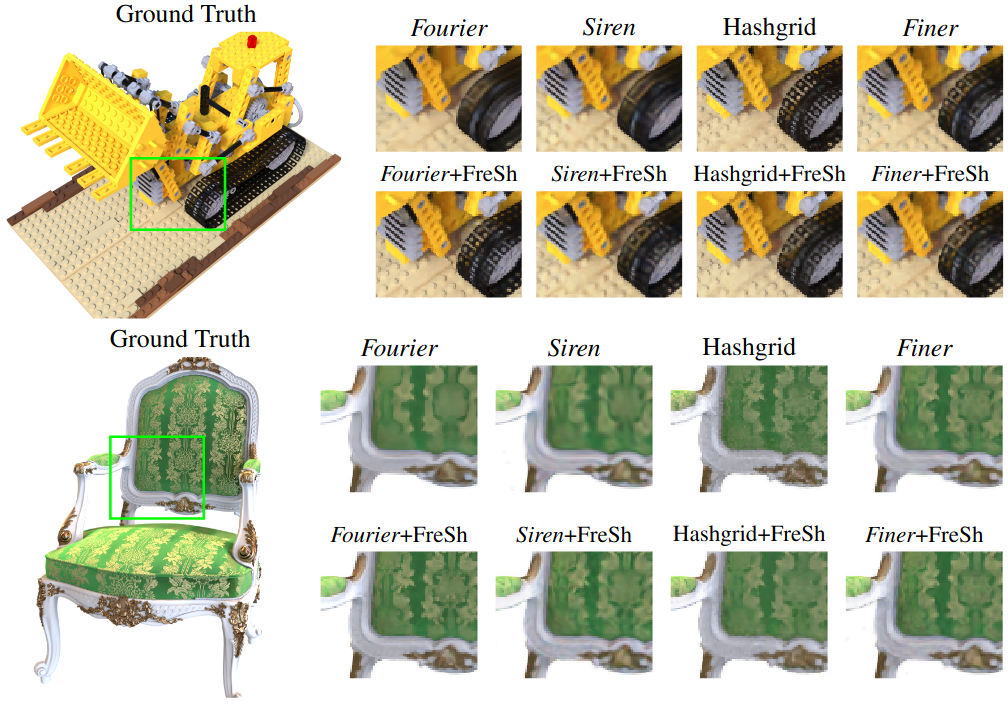}
    \centering
    \caption{Example model outputs for the object modeling task after approximately 20\% of the training.
    }
    \label{fig:nerf_examples_appendix}
\end{figure}

\paragraph{Object reconstruction} We provide additional examples in \Cref{fig:nerf_examples_appendix}.

\begin{table}[hb]
  \caption{Video representation results for NeRF, \siren{} and \ffeatures{} embeddings. \our{} outperforms baseline embedding configurations and NeRF. Each configuration was tested with and without time as an input coordinate. The model benefits from embeddings reconfigured with increased frequencies only when time is not an input, indicating that different frequency magnitudes are required for spatial and temporal directions. Results are averaged over 3 seeds.}
  \label{tab:video_fitting_full}
  \centering
  \begin{tabular}{@{\hspace{2mm}}l@{\hspace{2mm}}l@{\hspace{6mm}}l@{\hspace{4mm}}l@{\hspace{6mm}}l@{\hspace{4mm}}l@{\hspace{2mm}}}
        \toprule
        ~&Use & \multicolumn{2}{c}{Cat \qquad\qquad\qquad} &  \multicolumn{2}{c}{Bikes \qquad\qquad} \\ 
        ~&time  & PSNR & SSIM & PSNR & SSIM \\ 
        \midrule
        \siren{} & \cmark & $ 38.72 $ $\mathsmaller{\pm  0.05}$ &  $ 0.9518 $ $\mathsmaller{\pm  0.0004}$ & $ 41.12 $ $\mathsmaller{\pm  0.03}$ &  $ 0.9677 $ $\mathsmaller{\pm  0.0001}$ \\ 
        +\our{} & \cmark & $ 36.84 $ $\mathsmaller{\pm  0.03}$ & $ 0.9444$ $\mathsmaller{\pm 0.0002}$ & $ 40.28 $ $\mathsmaller{\pm  0.01}$ & $ 0.9684 $ $\mathsmaller{\pm  0.0002}$ \\  
        \siren{} & \xmark & $ 39.84 $ $\mathsmaller{\pm  0.03}$ & $ 0.9553 $ $\mathsmaller{\pm  0.0001}$ & $ 40.39 $ $\mathsmaller{\pm  0.02}$ & $ 0.9657 $ $\mathsmaller{\pm  0.0001}$ \\ 
        +\our{} & \xmark & $ \mathbf{40.61 }\mathsmaller{\pm  0.04}$ & $\mathbf{0.9579}\mathsmaller{\pm  0.0002}$ & $ \mathbf{41.62 }\mathsmaller{\pm  0.01}$ & $ \mathbf{0.9708 }\mathsmaller{\pm  0.0001}$ \\  
        \hline
        \ffeatures{} & \cmark & $ 38.14 $ $\mathsmaller{\pm  0.07}$ & $ 0.9481$ $\mathsmaller{\pm  0.0006}$ & $ 40.92 $ $\mathsmaller{\pm  0.07}$ & $ 0.9678$ $\mathsmaller{\pm  0.0006}$ \\  
        +\our{}  &  \cmark    & $37.38$ $\mathsmaller{\pm  0.04}$ & $0.9483$ $\mathsmaller{\pm  0.0002}$ &  $40.29$ $\mathsmaller{\pm  0.03}$  & $0.9696$ $\mathsmaller{\pm {>} 0.0001}$ \\  
        \ffeatures{} & \xmark & $ 38.82 $ $\mathsmaller{\pm  0.03}$ & $ 0.9514$ $\mathsmaller{\pm  0.0003}$ & $ 40.61$ $\mathsmaller{\pm  0.03}$ & $ 0.9667$ $\mathsmaller{\pm  0.0001}$ \\  
        +\our{}  &  \xmark & $\mathbf{39.68} \mathsmaller{\pm  0.04}$ & $\mathbf{0.9555 }\mathsmaller{\pm  0.0002}$ & $\mathbf{41.13 }\mathsmaller{\pm  0.02}$ & $\mathbf{0.9700 }\mathsmaller{\pm  0.0002}$ \\
        \hline
        NeRF & \cmark & $ \mathbf{37.39 }\mathsmaller{\pm  0.01}$ & $ \mathbf{0.9471 }\mathsmaller{\pm  0.0002}$ & $ \mathbf{39.78 }\mathsmaller{\pm  0.04}$ & $ \mathbf{0.9675 }\mathsmaller{\pm  0.0002}$ \\  
        NeRF & \xmark &  $ 37.13 $ $\mathsmaller{\pm  0.04}$ & $ 0.9462 $ $\mathsmaller{\pm  0.0002}$ & $ 39.57  $ $\mathsmaller{\pm  0.05}$  & $  0.9664$ $\mathsmaller{\pm  0.0004}$ \\ 
        \bottomrule
    \end{tabular}
\end{table}

\end{document}